%% file: iclr2027_conference.tex
\documentclass{article} % For LaTeX2e
\usepackage{iclr2027_conference,times}

\input{math_commands.tex}

\usepackage{graphicx}
\usepackage{amsmath}
\usepackage{booktabs}
\usepackage{pifont}   % \ding を使うため
\usepackage{amssymb}  % \blacktriangle を使うため
\usepackage{multirow} % これ忘れるな
\usepackage{soul}
\usepackage{ifthen}
\usepackage{etoolbox}
\usepackage{subcaption}
\usepackage{wrapfig}
\usepackage{adjustbox}

\usepackage{epigraph}
\usepackage{tcolorbox}
\usepackage{listings}
\usepackage{booktabs}
\usepackage{colortbl}
\usepackage{xcolor}
\definecolor{lightgreen}{RGB}{220,255,220}
\tcbuselibrary{skins}

\tcbset{
    promptbox/.style={
        colback=red!8!white,
        colframe=red!40!white,
        arc=6pt,
        boxrule=0.8pt,
        left=8pt, right=8pt, top=6pt, bottom=6pt,
        fonttitle=\small\bfseries\color{red!60!black},
        coltitle=red!60!black,
        attach boxed title to top left={yshift=-2mm, xshift=6mm},
        boxed title style={colback=red!8!white, colframe=red!40!white, arc=4pt, boxrule=0.8pt},
        fontupper=\small\ttfamily,
    }
}

\usepackage[utf8]{inputenc} % allow utf-8 input
\usepackage[T1]{fontenc}    % use 8-bit T1 fonts
\usepackage{hyperref}       % hyperlinks
\usepackage{url}            % simple URL typesetting
\usepackage{booktabs}       % professional-quality tables
\usepackage{amsfonts}       % blackboard math symbols
\usepackage{nicefrac}       % compact symbols for 1/2, etc.
\usepackage{microtype}      % microtypography
\usepackage{xcolor}         % colors
\usepackage{amsthm}
\newtheorem{theorem}{Theorem}
\newtheorem{lemma}{Lemma}
\newtheorem{proposition}{Proposition}
\newtheorem{corollary}{Corollary}
\newtheorem{assumption}{Assumption}
\newtheorem{remark}{Remark}

\usepackage{wrapfig}

\definecolor{promptbg}{HTML}{F8F9FA} % ほんのりグレー
\definecolor{promptframe}{HTML}{DEE2E6} % 控えめな枠線カラー
\definecolor{prompttext}{HTML}{212529} % 真っ黒ではないダークグレー

\lstdefinestyle{llmprompt}{
    backgroundcolor=\color{promptbg},
    rulecolor=\color{promptframe},
    basicstyle=\ttfamily\footnotesize\color{prompttext},
    frame=single,                   % 四方を枠線で囲む
    frameround=tttt,                % 枠線の角を少し丸くする（お好みで外してもOK）
    breaklines=true,                % 長い自然言語の折り返し（超重要）
    breakatwhitespace=true,         % 単語の途中で切れないようにする
    columns=fullflexible,           % 文字間隔を自然にする
    keepspaces=true,                % JSONなどのインデントを保持
    captionpos=b,                   % キャプションを下に配置
    xleftmargin=1em,                % 左右に少し余白を持たせる
    xrightmargin=1em,
    showstringspaces=false          % 文字列中のアンダースコアを非表示
}

\title{CyberWorld: World Models for Sample-Efficient Autonomous Cyber Defense}

\iclrfinalcopy
\author{%
Ryozo Masukawa$^{1}$,\\
Sanggeon Yun$^{1}$, 
Raheeb Hassan$^{1}$,
Hyunwoo Oh$^{1}$,
SungHeon Jeong$^{1}$,\\
Mohsen Imani$^{1}$\\[1ex]
$^{1}$University of California, Irvine\\
\texttt{rmasukaw@uci.edu}
}

\begin{document}

\maketitle
\input{sections/abstract}

\input{sections/intro}
\input{sections/background}

\input{sections/method}
\input{sections/experiments}
\input{sections/conclusion}

\bibliography{iclr2027_conference}
\bibliographystyle{iclr2027_conference}

\appendix
\section*{Appendix}
\renewcommand{\sectionautorefname}{Appendix}
\renewcommand{\subsectionautorefname}{Appendix}%
\renewcommand{\subsubsectionautorefname}{Appendix}%

\input{appendices/appendix_B_theory}
\input{appendices/appendix_A}

\input{appendices/appendix_C}

\end{document}

%% file: math_commands.tex
\usepackage{amsmath,amsfonts,bm}

\def\eqref#1{equation~\ref{#1}}
\def\1{\bm{1}}

\DeclareMathAlphabet{\mathsfit}{\encodingdefault}{\sfdefault}{m}{sl}
\SetMathAlphabet{\mathsfit}{bold}{\encodingdefault}{\sfdefault}{bx}{n}

%% file: sections/abstract.tex
\begin{abstract}
% Deep reinforcement learning is a promising approach to autonomous cyber defense, yet
% existing methods are predominantly model-free and require extensive environment
% interaction. World models have shown strong sample efficiency in robotics and embodied
% control by learning predictive dynamics and optimizing policies through imagined
% trajectories. Applying this paradigm to cybersecurity raises a fundamental question:
% what should constitute the ``world'' in a cyber world model? We introduce CyberWorld, a
% Dreamer-style framework for world-model-based cyber defense that learns latent cyber
% dynamics while supporting vector, graph, textual, and multimodal representations. On
% CyberWheel, we evaluate how representation choice affects sample efficiency, defensive
% reward, and robustness to network scale. CyberWorld exceeds a strategy-agnostic control
% defender after $3.6$k--$4.9$k environment steps, where PPO from scratch needs
% $2.3$M--$3.1$M or never succeeds: two to three orders of magnitude fewer interactions.
% Vector and graph representations are statistically tied, text is competitive at ten
% times the compute, and only graph succeeds on the topology-following attacker with every
% seed. From 15 to 100 hosts, the episodes needed to reach the control stay near 100, so
% the learned dynamics carry across scales. These results show the potential of world-
% model reinforcement learning for cyber defense and establish cyber-world representation
% as a central design problem.
Deep reinforcement learning has become a prominent approach to autonomous cyber
defense. Existing methods are predominantly model-free and consequently require
extensive environment interaction. World models provide an alternative by learning
predictive dynamics and optimizing policies through imagined trajectories, yielding
substantial gains in sample efficiency in robotics and embodied control. Extending this
paradigm to cybersecurity raises a fundamental question: what should constitute the
``world'' in a cyber world model? We introduce CyberWorld, a Dreamer-style world modeling framework
that learns latent cyber dynamics from vector, graph, textual, and multimodal
representations of the defended network. 
Across all four scoreable CyberWheel attack strategies, the graph-based
CyberWorld variant exceeds a strategy-agnostic control after $3.6$k--$15.8$k
environment steps, compared with millions of steps required by model-free PPO. 
Across representation choices, graph structure provides greater robustness under topology-dependent attacks, while simpler
representations remain competitive in overall performance. 
Among successful runs, the number of episodes required to reach the control remains approximately constant as network size increases from 15 to 100 hosts. These results establish learned cyber dynamics as a sample-efficient and scalable
basis for autonomous defense, and identify world representation as a central design axis
for robustness and scalability.
code is available \href{https://anonymous.4open.science/r/CyberWorld-768D/README.md}{here}\footnote{\href{https://anonymous.4open.science/r/CyberWorld-768D/README.md}{https://anonymous.4open.science/r/CyberWorld-768D/README.md}}.
\end{abstract}

%% file: sections/intro.tex
\section{Introduction}
\vspace{-3mm}

Reinforcement learning (RL)~\citep{sutton1998reinforcement} has emerged as a prominent approach to autonomous cyber defense, where agents make sequential decisions to protect networked systems against evolving threats~\citep{taxonomy_acd,kiely2025exploring}.
In autonomous cyber defense, it is common to rely on \emph{model-free} RL algorithms~\citep{ppo,dutta2023deep,hmarl,cybermonic,kazeminajafabadi2026posterior}, directly optimizing defensive policies through repeated environment interaction.
However, this prevailing paradigm is increasingly inadequate in the face of modern offensive capabilities.
With frontier agentic models such as Claude Mythos~\citep{anthropicClaudeMythos}, autonomous adversaries can increasingly reason over complex systems and exploit zero-day vulnerabilities~\citep{zero_d}, intensifying the challenge faced by purely reactive DRL defenders.
Recent evidence demonstrates this fragility. 
\citet{masukawa2026trident} showed that even compact agentic language models~\citep{slm_future} can autonomously discover and exploit blind spots in state-of-the-art DRL blue agents. 
Separately, \citet{sok_usenix} systematized the pitfalls of RL-based cyber defense, finding that existing defenses frequently fail to account for non-stationary adversaries, inadequately address partial observability, and lack demonstrated policy convergence.
Beyond these robustness issues, model-free RL can also be sample-inefficient, requiring substantial environment interaction for learning and adaptation~\citep{aamas_mamba}.

Meanwhile, world models have advanced rapidly in robotics, visual control, and embodied intelligence~\citep{dreamer_v3,zhang2023storm,alonso2024diffusion}.
These methods learn predictive environment dynamics and optimize decisions through imagined trajectories~\citep{dreamer_v3,iris}.
This separation between dynamics modeling and policy learning is particularly attractive in cyber defense because the two components change at different rates. An attacker may change its strategy from one campaign to the next, but the network protocols, services, and detection mechanisms through which both attacker and defender act, that is, the world itself, do not~\citep{cyberwheel,kiely2025exploring}. A model of these shared dynamics therefore remains valid when the adversary changes, whereas a reactive policy fitted to one adversary must be relearned.
\begin{wrapfigure}{r}{0.68\textwidth}
    \centering
    \includegraphics[width=\linewidth]{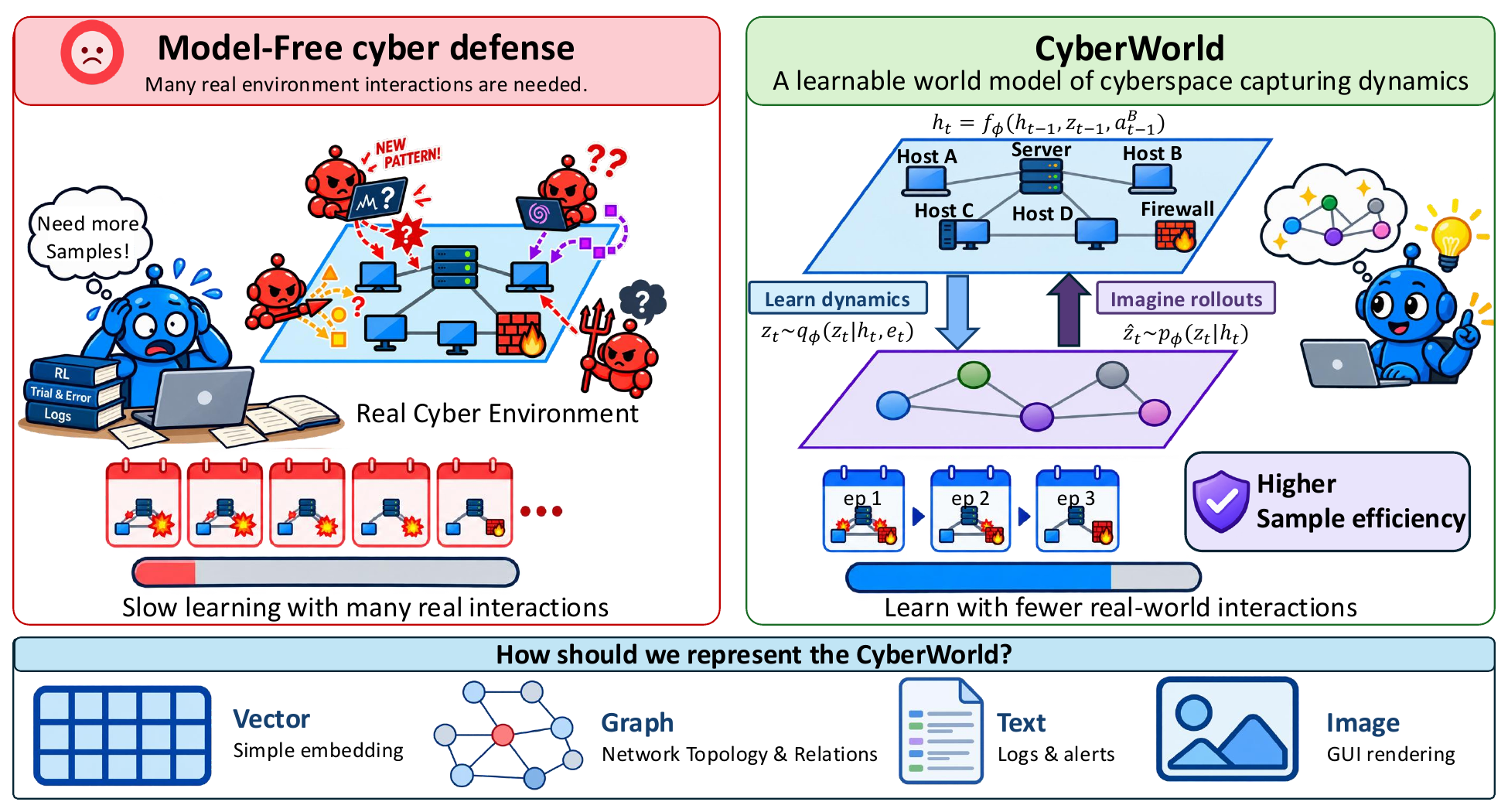}
    \vspace{-7mm}
    \caption{CyberWorld learns predictive cyber dynamics to enable more sample-efficient autonomous defense.}
    \label{fig:placeholder}
    \vspace{-6mm}
\end{wrapfigure}
Recent work further extends world models to visual, language, and multimodal representations~\citep{zhou2025dinowm,lin2024learning,mazzaglia2024genrl}, which suggests that cyber dynamics, too, may be learnable from heterogeneous observations and that a learned model may enable substantially more sample-efficient autonomous defense.

Transferring world models to cybersecurity, however, first requires deciding what the cyber ``world'' is. Unlike visual control, where observations take standardized forms such as pixels or proprioceptive signals~\citep{dreamer_v3,zhou2025dinowm}, a cyber defender observes heterogeneous information, including host attributes, network topology, alerts, logs, and their temporal histories, and the suitability of a given representation may change with the scale and structure of the network. The central question of this work is therefore: \textbf{What should constitute the ``world'' in a world model for autonomous cyber defense?}

To investigate this question, we introduce \textbf{CyberWorld}, a Dreamer-style~\citep{dreamer_v1,dreamer_v2,dreamer_v3} world model for autonomous cyber defense.
CyberWorld learns latent cyber dynamics and trains defensive policies through imagined trajectories.
Within a shared world-model framework, we study five representations of cyberspace: per-host vectors, host graphs, their combination, natural-language telemetry, and vector--text fusion with an added frame-reconstruction objective.

Using CyberWheel~\citep{cyberwheel}, we evaluate three aspects of CyberWorld: how cyber representations affect world-model-based control, how they scale as the network grows from 15 to 100 hosts, and whether world-model-based defense can substantially improve sample efficiency over model-free PPO~\citep{ppo}.
We also examine end-of-training reward stability to distinguish fast initial learning from stable defensive behavior.

Our experiments demonstrate a substantial reduction in environment interaction.
Across all four scoreable attack strategies, the graph-based CyberWorld
variant reaches the control within $3.6$k--$15.8$k steps, while PPO requires
$2.3$M--$3.1$M steps or fails within its $3.2$M-step budget.
Vector and graph representations perform comparably on most attackers, and the
graph representation is the most reliable on the topology-following attacker.
On larger networks, successful runs require a similar number of episodes to reach
the control, although the fixed interaction budget becomes restrictive at 100 hosts.
End-of-training reward stability varies across representations, indicating that rapid
acquisition of a defensive behavior does not by itself imply convergence.
% Our experiments demonstrate a substantial reduction in environment interaction. On three of the four attack strategies, CyberWorld exceeds a strategy-agnostic control after $3.6$k--$4.9$k environment steps, whereas PPO requires $2.3$M--$3.1$M steps or fails within its $3.2$M-step budget, a reduction of $470$--$850\times$. Vector and graph representations perform comparably on most attackers, and the graph representation is the most reliable on the topology-following attacker. On larger networks, successful runs require a similar number of episodes to reach the control, although the fixed interaction budget becomes restrictive at 100 hosts. End-of-training reward stability varies across representations, indicating that rapid acquisition of a defensive behavior does not by itself imply convergence.

Our contributions are threefold:
\begin{itemize}
    \item We introduce \textbf{CyberWorld}, to the best of our knowledge, the first Dreamer-style world-model framework for autonomous cyber defense, learning predictive cyber dynamics and training defensive policies through imagined trajectories.
    
    \item We investigate which representation of the cyber world is appropriate for world modeling by comparing various representations, and further study the vector and graph representations across network scales; the results highlight the benefit of explicit graph structure when attack progression depends on network topology.    
    
    \item We demonstrate that CyberWorld can achieve effective cyber defense with orders of magnitude fewer environment interactions than model-free PPO.
\end{itemize}

%% file: sections/background.tex
\vspace{-3mm}
\section{Background \& Related Works}
\vspace{-3mm}
\subsection{From Model-Free Reinforcement Learning to Multimodal World Models}

RL formulates sequential decision making through interactions between an agent and its environment \citep{sutton1998reinforcement}. Modern deep RL has been dominated by model-free approaches such as PPO \citep{ppo}, which optimize policies directly from collected experience without explicitly learning environment dynamics. While effective, this paradigm can require substantial environment interaction, motivating renewed interest in model-based learning for improved sample efficiency. 
PlaNet~\citep{planet} showed that latent dynamics learned from high-dimensional observations can support planning. Building on this, the Dreamer family further established latent imagination as a practical framework for learning policies primarily inside a predictive world model \citep{dreamer_v1,dreamer_v2,dreamer_v3}. Recent work has expanded this framework with Transformer-based dynamics and more accurate latent prediction \citep{micheli2023transformers,robine2023transformerbased,emerald}. At the same time, world models have begun to move beyond primarily visual observations: \citet{lin2024learning} jointly model language and vision for imagined policy learning, while \citet{mazzaglia2024genrl} connect multimodal foundation representations with latent environment dynamics. 
Structured representations have also emerged as an alternative to monolithic latent states, including causal graph formulations for explicit state reasoning \citep{zhou2026speech}. These developments motivate CyberWorld's central question of how heterogeneous cyber observations, including vectors, network topology, text, and visual interfaces, should be represented within a predictive world model.

\vspace{-3mm}
\subsection{Model-Based Autonomous Cyber Defense}
\vspace{-3mm}

Model-based cyber defense has traditionally relied on environment dynamics that are manually specified or explicitly estimated rather than learned as general predictive latent models. Prior work has used transition, observation, or causal models to support belief tracking, planning, and defensive decision making~\citep{pomdp_model,cyber_18,andrew2022developing}, including domain-specific Partially Observable Markov Decision Process (POMDP) dynamics updated online from observations~\citep{acd_multi_strategy}. In contrast, most recent DRL approaches learn defensive policies directly from interaction without explicitly modeling how the cyber environment evolves~\citep{dutta2023deep}. Recent work has begun to bridge these two paradigms by learning selected aspects of the environment. \citet{kazeminajafabadi2026posterior}, for example, infer latent attacker context to condition defensive policies while leaving the underlying environment dynamics implicit; latent dynamics have also been used for MCTS-based cyber-defense planning~\citep{li2026acdzero}. Closest to CyberWorld, \citet{karacelebi2026learning} learn a graph-structured world model for networked systems with recurrent latent dynamics and imagined rollouts. Their setting, however, models non-adversarial network evolution. To the best of our knowledge, CyberWorld is the first Dreamer-style world model for autonomous cyber defense, learning predictive latent dynamics that capture the interaction between attacker behavior, defender interventions, and evolving network state.

%% file: sections/method.tex
\vspace{-3mm}
\section{CyberWorld}
\vspace{-3mm}
\label{sec:method}

\begin{wrapfigure}{r}{0.56\textwidth}
    \vspace{-3mm}
    \centering
    \includegraphics[width=\linewidth]{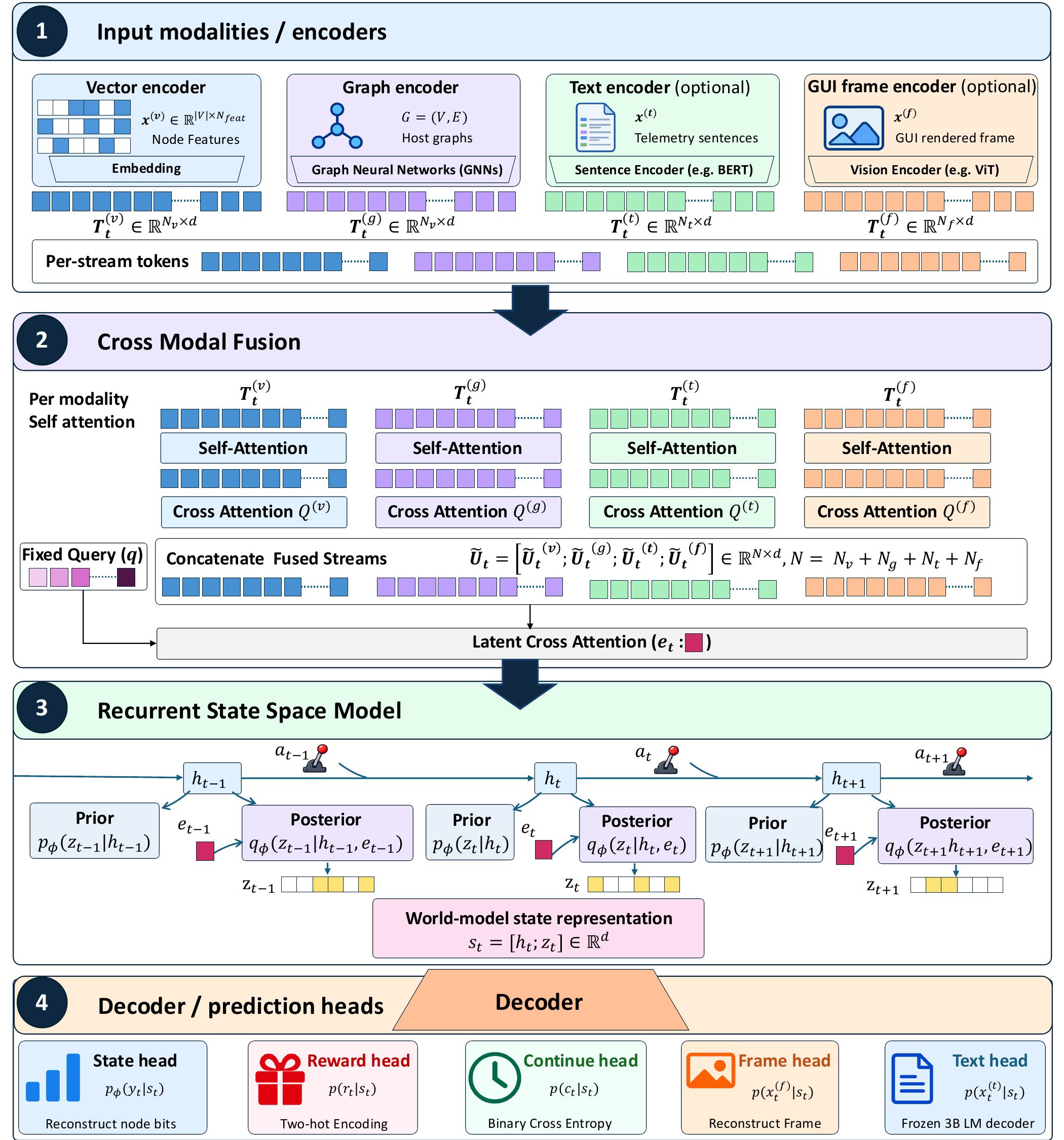}
    \caption{Architecture of CyberWorld. Multimodal telemetry and structural observations are mapped into a unified latent space via cross-attention pooling, driving an RSSM that tracks recurrent belief dynamics and trains defensive behavior purely within imagined rollouts.}
    \label{fig:cyberworld_detail}
    \vspace{-5mm}
\end{wrapfigure}

CyberWorld is a multimodal latent world model tailored for autonomous cyber defense, built upon the DreamerV3 framework~\citep{dreamer_v3} and adapted to the discrete, sparse-reward, and relational topology of enterprise network defense. The agent constructs a predictive model of the environment from replayed interactions and trains its policy entirely inside the model's imagination, decoupling the sample complexity of behavior learning from the high computational overhead of interacting with live network simulators. {Its cyber-specific components are an environment-independent intermediate representation of the defender's view, encoders for heterogeneous cyber observations (host vectors, the host graph, and telemetry text) that are fused into a single latent embedding, and modality-specific reconstruction objectives that determine what the latent state must preserve (\autoref{fig:cyberworld_detail}).}

\vspace{-3mm}
\subsection{Problem Formulation}
\vspace{-3mm}
\label{sec:setting}

We instantiate our formulation in CyberWheel~\citep{cyberwheel}, a configurable simulation environment for autonomous cyber defense. CyberWheel models enterprise networks with explicit hosts, subnets, routing, services, and firewall constraints, while scripted adversaries progress through MITRE ATT\&CK- and Atomic Red Team-inspired attack stages~\citep{mitreMITREATTampCKxAE,atomicredteamAtomicTeam}. The defender receives only detector alerts and the state of its own assets, and responds through cyber deception by deploying or removing decoy hosts along the adversary's likely path. Compared with other cyber RL environments~\citep{msft:cyberbattlesim,NASimEmu,kiely2025exploring}, CyberWheel is particularly well suited to controlled world-model evaluation because it combines a centralized single-defender setting with configurable attacker strategies, detector models, and network topologies. This design avoids additional confounds from multi-agent coordination and credit assignment~\citep{victim_marl,aamas_mamba,zhang2025combo,xue2026learning,zhao2026empowering}, while allowing the same simulation substrate to support our representation and scalability studies.

We formalize blue-agent network defense as a POMDP $(\mathcal{S}, \mathcal{A}^B, \mathcal{O}, \mathcal{T}, \Omega, R, \gamma)$, where $\mathcal{S}$ represents the underlying global environment states (network topology, host compromise status, decoy configurations, and latent adversarial intent), $\mathcal{A}^B$ denotes the defender's action set, $\mathcal{O}$ is the observation space, $\mathcal{T}$ specifies the global transition dynamics, $\Omega(o \mid s)$ maps states to partial defender observations, $R(s, a^B)$ is the reward function, and $\gamma \in [0,1)$ is the temporal discount factor.

In enterprise network defense, the transition dynamics $\mathcal{T}$ govern the coupled execution of defender countermeasures and adversarial tactics. Let $a_t^B \in \mathcal{A}^B$ denote the blue defender action, $a_t^R \in \mathcal{A}^R$ denote the concurrent red adversary action, and $c \in \mathcal{C}$ represent an adversary strategy. From the defender's perspective, the effective state transition $\mathcal{T}_c(s_{t+1} \mid s_t, a_t^B)$ factors into:
\begin{equation}
\mathcal{T}_c(s_{t+1} \mid s_t, a_t^B)
=
\sum_{a_t^R \in \mathcal{A}^R}
\mathcal{T}_{\mathrm{mech}}(s_{t+1} \mid s_t, a_t^B, a_t^R)
\, \pi_c^R(a_t^R \mid b_t^R),
\label{eq:transition_factorization}
\end{equation}
where $\mathcal{T}_{\mathrm{mech}}$ captures the underlying \emph{cyber mechanisms} (protocol semantics, service configurations, vulnerability exploits, and deception capture constraints), which constrain the attacker as much as the defender, since an adversary can exploit a host only through the protocols and services the network exposes, while $\pi_c^R$ reflects the adversary's stochastic policy conditioned on their internal belief $b_t^R$. When an adversary shifts tactics ($c \to c'$), $\pi_c^R$ changes, inducing an apparent shift in overall dynamics $\mathcal{T}_c \to \mathcal{T}_{c'}$. However, $\mathcal{T}_{\mathrm{mech}}$ is shared across strategies under this formulation. A model-free policy does not explicitly preserve this shared transition factor, whereas a latent world model separates dynamics prediction from policy learning and can, in principle, retain mechanism information across changes in adversary strategy.
We provide a formal analysis of this factorization in \autoref{app:theory}, showing how adversary-strategy shifts affect the effective transition kernel and how model error decomposes into shared-mechanism and strategy-dependent components.

The defender's observation $o_t$ is the stream of detector alerts and asset states described above, which we refer to as \emph{telemetry}. Telemetry is incomplete and delayed and never exposes the adversary's internal state, so the observation sequence is non-Markovian. Optimal defense therefore requires a temporal belief state $b_t(s) = p(s_t = s \mid o_{\le t}, a_{<t}^B)$, which CyberWorld approximates by a recurrent latent state $s_t = [\,h_t;\, z_t\,] \approx b_t(s)$. Under this formulation, the choice of observation representation $\psi(o_t)$ is a dynamics-modeling decision: a representation is adequate when the sequence $\psi(o_{\le t})$ carries sufficient statistics to identify $\mathcal{T}_{\mathrm{mech}}$ and to track $b_t^R$.
Because attacks manifest both as localized changes in host status and as non-local traversals of the network topology, a representation must capture both local state and topological information for the transitions under $\mathcal{T}_c$ to be predictable.
% Because attacks manifest both as localized changes of host status and as non-local traversals of the topology, a representation must expose both kinds of information for the transitions under $\mathcal{T}_c$ to be predictable.

\subsection{World Model}
\label{sec:wm}

{The world model follows the data flow of \autoref{fig:cyberworld_detail}: each observation channel is encoded into a token sequence, the token sequences are fused into one observation embedding $e_t$, a Recurrent State-Space Model (RSSM)~\citep{planet, dreamer_v3} turns the sequence of embeddings and defender actions into a latent state $s_t$, and modality-specific heads reconstruct the observation, the reward, and the continuation flag from $s_t$. Let $\mathcal{M}$ denote the set of channels an arm encodes and $\mathcal{K}$ the set of heads it trains; both are configuration, and every arm shares the same fusion, latent dynamics, head architectures, and behavior learning.}

% \paragraph{Modality Encoders.}
% Each channel $m \in \mathcal{M}$ produces a token sequence
% $T_t^m \in \mathbb{R}^{L_m \times d}$ of common width $d$.
% Vector observations are encoded independently per host, while the graph
% encoder augments the same host features with graph-attention message passing
% over $G$~\citep{gat}. Text telemetry $w_t$ is encoded by a frozen pretrained
% language model~\citep{qwen2.5} and projected into the shared token space through
% a trainable adapter. Visual observations $I_t$ are similarly mapped into
% patch-level tokens by a visual encoder and projected into the shared token
% space; when enabled as a prediction target, they are reconstructed by the
% corresponding frame head. Frame encoding is optional and unused in the reported configurations.
\paragraph{Modality Encoders.}
Each channel $m \in \mathcal{M}$ produces a token sequence
$T_t^m \in \mathbb{R}^{L_m \times d}$ of common width $d$.
Vector observations are encoded independently per host, while the graph
encoder augments the same host features with graph-attention message passing
over $G$~\citep{gat}. Text telemetry $w_t$ is encoded by a frozen pretrained
language model~\citep{qwen2.5} and projected into the shared token space through
a trainable adapter. Visual observations $I_t$ are similarly mapped into
patch-level tokens by a visual encoder and projected into the shared token
space; when enabled as a prediction target, they are reconstructed by the
corresponding frame head. Visual encoding is optional and unused in the reported configurations.

\paragraph{Observation Representations.}
\label{sec:preprocessing}
Every channel is rendered from one environment-agnostic intermediate representation (IR) of the defender's view, which is the only object an encoder or renderer receives; ground-truth compromise labels and the adversary's position are kept in a separate annotation record used for analysis only. The IR fixes the node set before an episode starts: the $n_{\mathrm{host}}$ physical hosts occupy the indices given by the network configuration, and $n_{\mathrm{decoy}}$ reserved per-subnet slots receive decoys as they are deployed, each decoy binding to the lowest free slot of its subnet and releasing it on removal, so that the $N = n_{\mathrm{host}} + n_{\mathrm{decoy}}$ node indices refer to the same hosts throughout an episode and never reveal decoys that have not yet been deployed. The vector channel is the defender-visible host state
{\footnotesize
\begin{equation}
\mathbf{x}_t \in \{0,1\}^{N \times 5},
\qquad
\mathbf{x}_t[i,:] = \big(\texttt{present},\, \texttt{is\_decoy},\, \texttt{isolated},\, \texttt{alert\_now},\, \texttt{alert\_ever}\big)_i,
\end{equation}
}
whose alert fields originate from the defender's own detectors~\citep{taxonomy_acd} and are keyed on the source host; the IR carries neither a destination host nor a technique label, since either would reveal the attacker's next target. For reconstruction, the five fields are mapped to four role bits
{\footnotesize
\begin{equation}
\bm{y}_t[i,:]
=
\big(
\max(\texttt{present},\,\texttt{isolated}),\,
\texttt{is\_decoy},\,
\texttt{alert\_now},\,
\texttt{alert\_ever}
\big)_i
\in \{0,1\}^{4}.
\label{eq:rolebits}
\end{equation}
}
The encoder uses $\bm{x}_t$, while the state head reconstructs $\bm{y}_t$.
The graph channel adds the sparse adjacency $G = (V, E)$ with $|V| = N$, taken from the network configuration: hosts on the same subnet are adjacent, cross-subnet edges exist where an interface is declared, and the topology is static within an episode. The text channel $w_t$ renders the same record as prose, namely an alert summary, the defender's own action, the source host of any detected activity, the changes since the previous step, and the present hosts grouped by alert state with decoys named individually. The frame $I_t$ draws the nodes at positions determined by the topology, adds the edges of $G$, and colours each present node by $\bm{y}_t$, so a frame is exactly invertible to the role bits. All channels are deterministic functions of the same IR record; every arm therefore receives the same information, and the arms differ in representation alone. Benchmark-specific constants are given in \autoref{sec:exp_setup}.

% whose alert fields originate from the defender's own detectors~\citep{taxonomy_acd} and are keyed on the source host; the IR carries neither a destination host nor a technique label, since either would reveal the attacker's next target. 
% The five fields are grouped into four roles (availability, defender asset, activity, and alert history) on which the renderers and the reconstruction target operate, so a new environment is added by declaring a field-to-role mapping. The graph channel adds the sparse adjacency $G = (V, E)$ with $|V| = N$, taken from the network configuration: hosts on the same subnet are adjacent, cross-subnet edges exist where an interface is declared, and the topology is static within an episode. The text channel $w_t$ renders the same record as prose, namely an alert summary, the defender's own action, the source host of any detected activity, the changes since the previous step, and the present hosts grouped by alert state with decoys named individually. The frame $I_t$ draws the nodes at positions determined by the topology, adds the edges of $G$, and colours each present node by its four role bits, so a frame is exactly invertible to the role bits. All channels are deterministic functions of the same IR record; every arm therefore receives the same information, and the arms differ in representation alone. Benchmark-specific constants are given in \autoref{sec:exp_setup}.

\paragraph{Cross-Modal Fusion.}
Fusion keeps the channels as separate token streams until a final pooling step. Writing $\operatorname{Attn}(Q, K, V) = \operatorname{softmax}\!\big(QK^{\top}/\sqrt{d}\big)\,V$ for multi-head attention, each stream is first tagged with a learned channel embedding and refined by a stack of self-attention blocks in which the stream attends to itself,
\begin{equation}
U^m_t = \operatorname{Attn}\big(T^m_t,\; T^m_t,\; T^m_t\big),
\label{eq:selffusion}
\end{equation}
and, when more than one channel is active, each refined stream then attends to the concatenation of the other streams,
\begin{equation}
\tilde{U}^m_t = \operatorname{Attn}\big(U^m_t,\; [\,U^{m'}_t\,]_{m' \neq m},\; [\,U^{m'}_t\,]_{m' \neq m}\big),
\label{eq:crossfusion}
\end{equation}
so that, for instance, a text token can attend to the node token it describes. A single learned query $q \in \mathbb{R}^d$ finally pools the concatenated fused streams into a fixed-size observation embedding~\citep{jaegle2021perceiver},
\begin{equation}
e_t = \operatorname{Attn}\big(q,\; [\,\tilde{U}^m_t\,]_{m \in \mathcal{M}},\; [\,\tilde{U}^m_t\,]_{m \in \mathcal{M}}\big) \in \mathbb{R}^d,
\label{eq:fusion}
\end{equation}
whose size is independent of the number of hosts or text tokens. With a single active channel, the cross-attention stage is omitted and $q$ pools $U^m_t$ directly.

\paragraph{Latent Dynamics.}
{The RSSM maintains a deterministic recurrent state $h_t$ and a discrete stochastic state $z_t$; with parameters $\phi$,}
\begin{equation}
\begin{cases}
\text{Recurrent model:} & h_t = f_\phi(h_{t-1}, z_{t-1}, a_{t-1}^B) \\[2pt]
\text{Posterior representation:} & z_t \sim q_\phi(z_t \mid h_t, e_t) \\[2pt]
\text{Prior transition dynamics:} & \hat{z}_t \sim p_\phi(\hat{z}_t \mid h_t),
\end{cases}
\label{eq:rssm}
\end{equation}
{where the posterior incorporates the current fused observation $e_t$, while the prior predicts the stochastic state from history alone and is used for imagined rollouts. The remaining stochastic parameterization follows DreamerV3~\citep{dreamer_v3}.}

The concatenated latent vector
\begin{equation}
s_t = [\,h_t;\, z_t\,]
\end{equation}
constitutes the model state supplied to all decoding heads, policy actors, and value critics{; $s_t$ is the learned counterpart of the defender belief $b_t$ and only the defender's own action conditions it}.

\paragraph{{Prediction and Reconstruction Heads.}}
\label{sec:wm-heads}

{Every head reads the posterior state $s_t$ and reconstructs a quantity of the same step $t$: the model is trained to explain the present from its belief, and the future enters only through the prior in \autoref{eq:rssm}.}
All configurations reconstruct the four defender-visible node-state roles $\bm{y}_t$ with a shared per-node decoder, and predict reward and episode continuation. The state decoder conditions on $s_t$ and a learned node query, so its parameters are independent of network size. The reward head uses distributional symlog two-hot parameterization~\citep{dreamer_v3}. Text-enabled configurations additionally reconstruct $w_t$ through the frozen language model conditioned on a learned latent prefix, while the multimodal configuration reconstructs $I_t$ with a convolutional decoder.

% \paragraph{{Prediction and Reconstruction Heads.}}
% \label{sec:wm-heads}

% {Every head reads the posterior state $s_t$ and reconstructs a quantity of the same step $t$: the model is trained to explain the present from its belief, and the future enters only through the prior in \autoref{eq:rssm}.}
% All configurations reconstruct the four defender-visible node-state roles $\bar{x}_t$ with a shared per-node decoder, and predict reward and episode continuation. The state decoder conditions on $s_t$ and a learned node query, so its parameters are independent of network size. The reward head uses distributional symlog two-hot parameterization~\citep{dreamer_v3}. Text-enabled configurations additionally reconstruct $w_t$ through the frozen language model conditioned on a learned latent prefix, while the multimodal configuration reconstructs $I_t$ with a convolutional decoder.

\paragraph{{Cyber-Specific Loss Weighting.}}
{Two properties of cyber telemetry make the standard reconstruction objectives degenerate, and CyberWorld modifies them accordingly; both are departures from DreamerV3.}

{First, most node attributes persist between consecutive steps, allowing an unweighted objective to succeed largely by predicting no change. We therefore up-weight state elements that change:}
\begin{equation}
\mathcal{L}_{\mathrm{state}} = \frac{1}{4N}\sum_{i=1}^{N}\sum_{j=1}^{4}
\lambda_{t,i,j}\,\operatorname{BCE}\big(
\hat{y}_{t,i,j},\, y_{t,i,j}
\big),
\qquad
\lambda_{t,i,j} =
\begin{cases}
\rho_x & \text{if } y_{t,i,j} \neq y_{t-1,i,j} \\
1 & \text{otherwise.}
\end{cases}
\label{eq:changeweight}
\end{equation}
% \begin{equation}
% \mathcal{L}_{\mathrm{state}} = \frac{1}{4N}\sum_{i=1}^{N}\sum_{j=1}^{4} \lambda_{t,i,j}\,\operatorname{BCE}\big(\hat{x}_{t,i,j},\, x_{t,i,j}\big),
% \qquad
% \lambda_{t,i,j} =
% \begin{cases}
% \rho_x & \text{if } x_{t,i,j} \neq x_{t-1,i,j} \\
% 1 & \text{otherwise.}
% \end{cases}
% \label{eq:changeweight}
% \end{equation}

Second, the reward is dominated by rare events, namely the decoy captures and decoy-budget violations, which occur on a small fraction of steps and therefore contribute little to a uniformly weighted reward loss. CyberWorld balances the reward-bearing steps $\mathcal{R}=\{t:|r_t|\geq\tau\}$ and the remaining steps $\mathcal{R}^{c}$ within each batch:
\begin{equation}
\mathcal{L}_{\mathrm{reward}} = \sum_{t} \omega_t\, \ell_t,
\qquad
\omega_t =
\begin{cases}
\tfrac{1}{2|\mathcal{R}|} & \text{if } t \in \mathcal{R} \\[2pt]
\tfrac{1}{2|\mathcal{R}^{c}|} & \text{otherwise,}
\end{cases}
\label{eq:rewardbalance}
\end{equation}
{where $\ell_t$ is the two-hot reward loss. When a batch contains one group, the mean is used.}

\paragraph{World-Model Objective.}
{Over replayed trajectory segments, the world model parameters $\phi$ are trained to minimise
\begin{equation}
\mathcal{L}(\phi) = \mathbb{E}_{q_\phi}\Big[\sum_{t}\Big(\sum_{k \in \mathcal{K}} \beta_k\, \mathcal{L}_k + \beta_{\mathrm{dyn}}\,\mathcal{L}_{\mathrm{dyn}} + \beta_{\mathrm{rep}}\,\mathcal{L}_{\mathrm{rep}}\Big)\Big],
\label{eq:wmloss}
\end{equation}
}
{where $\mathcal{L}_k$ denotes the active prediction and reconstruction losses, and $\mathcal{L}_{\mathrm{dyn}}$ and $\mathcal{L}_{\mathrm{rep}}$ are DreamerV3's dynamics and representation KL objectives~\citep{dreamer_v3}.}

\vspace{-2mm}
\subsection{Behavior Learning and Online Training}
\vspace{-2mm}
\label{sec:ac}
\label{sec:loop}

The actor and the critic are trained entirely on trajectories imagined by the world model, and both follow DreamerV3 unchanged~\citep{dreamer_v3}: imagined trajectories start from posterior states of replayed segments and are rolled forward through the RSSM prior, and the actor and the distributional critic are optimized with bootstrapped $\lambda$-returns, REINFORCE, entropy regularization, and percentile-based return normalization. Training likewise follows DreamerV3's concurrent scheme. The agent collects one episode with its current policy, appends it to a growing replay buffer, and performs a fixed number of world-model and actor--critic updates on sampled segments before collecting the next episode. Training starts from randomly initialized parameters with no offline data, and every environment interaction, including the random prefill of the buffer, counts toward the reported budget; the prefill size, update ratio, and budget are given in \autoref{sec:exp_setup}. The one cyber-specific consideration is that every declared defender action remains available at every step, so the policy can avoid the decoy-budget penalty only if the world model represents it, which is what the balanced reward objective of \autoref{eq:rewardbalance} provides.

%% file: sections/experiments.tex
\vspace{-5mm}
\section{Experiments}
\vspace{-3mm}
\input{tables_w_gnn_ppo/T2_scale}

\vspace{-2mm}
\subsection{Experimental Setup}
\vspace{-2mm}

\label{sec:exp_setup}

Our evaluation answers three Research Questions (RQs):
(\textbf{RQ1}) Can a world model substantially improve the sample efficiency
of autonomous cyber defense over model-free DRL?
(\textbf{RQ2}) Which representation of the cyber environment best supports
world-model-based control?
(\textbf{RQ3}) Does the answer survive changes in network scale?
RQ1 establishes whether learning the dynamics pays off in the resource that is
scarce when a defender is trained against a live or emulated network, the number of
environment interactions. RQ2 determines which of the representations of
\autoref{sec:preprocessing} lets the world model capture the dynamics that matter for
control, and RQ3 whether that answer and the interaction saving hold as the defended
network grows.

\noindent\textbf{Environment.} All experiments run in CyberWheel~\citep{cyberwheel}
against seven scripted ART attacker strategies, four of which are scoreable (the other
three have a control return of 0). RQ1 and RQ2 use the 15-host, three-subnet network;
RQ3 scales the same network to $N \in \{25, 50, 100\}$ hosts with the decoy budget
scaled with $N$. Each CyberWorld and GNN-PPO run uses the same 52{,}500-step interaction
budget, corresponding to 1{,}050 episodes at $N{=}15$ (50 random prefill and 1{,}000 training episodes).

\noindent\textbf{Representation arms.} Five arms share the latent dynamics, behavior
learning, and hyperparameters and differ only in encoders and reconstruction heads:
\emph{vector} (per-node MLP + self-attention), \emph{graph} (GAT over the host graph),
\emph{vec+graph}, \emph{text} (frozen Qwen2.5-3B over telemetry sentences), and
\emph{all} (vector + text fusion with an added frame-reconstruction
head). All share one training configuration: DreamerV3's return normalisation, the balanced reward loss of \autoref{eq:rewardbalance}, a 1{,}000-step actor--critic warm-up, and cosine learning-rate decay.

\noindent\textbf{Baselines.} \emph{PPO} is CyberWheel's own PPO trained from scratch
under the same environment and observation but with its original substantially
longer training budget, reaching up to 3.2M environment steps at $N{=}15$.
\emph{GNN-PPO} is a representation-matched baseline: the graph arm's observation and
GAT encoder trained with PPO under the same online interaction budget and seeds
as CyberWorld, without a world model. 
The strategy-agnostic \texttt{deploy\_then\_stop} policy serves as a fixed
rule-based control, defining the minimum return threshold that a learned
defender must exceed; the environment reward is unchanged throughout.
% The strategy-agnostic \texttt{deploy\_then\_stop} policy is the control every method must outperform; the environment reward is unchanged throughout.

\noindent\textbf{Metrics.}
The primary metric is the number of environment steps until the trailing-10 episode return first exceeds the control, reported as mean $\pm$ sd over 3 seeds; four text/multimodal cells use 2 seeds because the third run did not complete. Runs that never cross are reported as such. Curves show the trailing-10 return and a trailing-50 trend. PPO's crossing uses its 30-episode checkpoint evaluations, ours the online training return; no accuracy-based metric is used anywhere.

\begin{figure}[t]
    \centering
    \includegraphics[width=\linewidth]{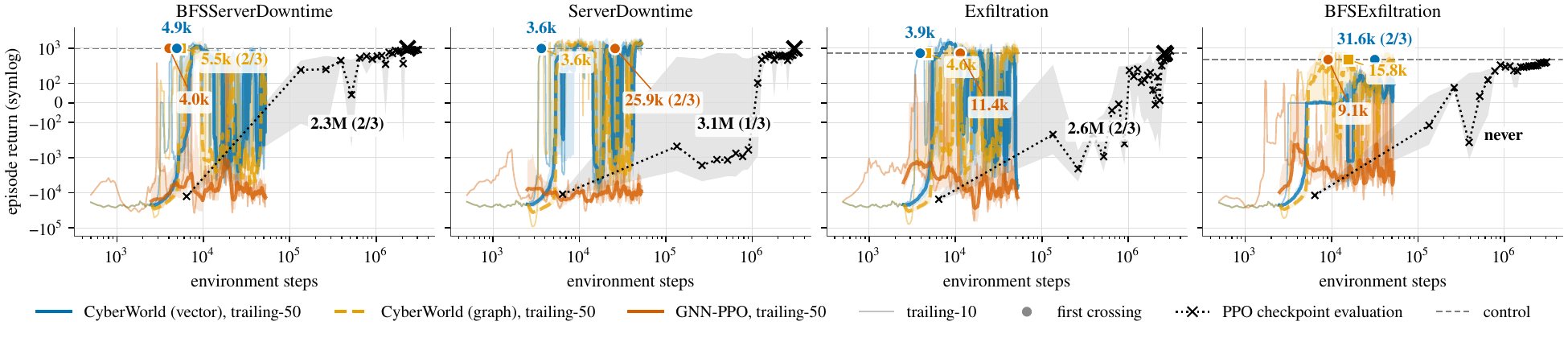}
    \vspace{-6mm}
    \caption{Return vs.\ environment steps for CyberWorld (vector and graph), GNN-PPO, and PPO on the four
    scoreable attackers at $N{=}15$ (seed mean, band = min/max over three seeds; PPO:
    30-episode checkpoint evaluations). Dashed: the \texttt{deploy\_then\_stop} control;
    markers: mean first crossing, with the seed count where not all seeds crossed.}
    \vspace{-3mm}
    \label{fig:rq1}
\end{figure}

\begin{figure}[t]
    \centering
    \includegraphics[width=0.94\linewidth]{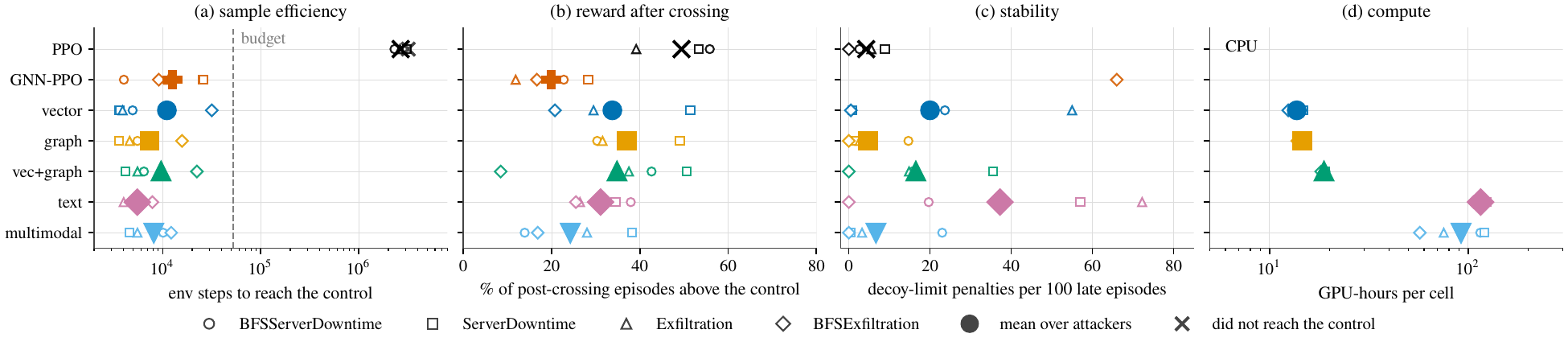}
    \vspace{-3mm}
    \caption{Representation comparison at $N{=}15$: (a) steps to control, (b) post-crossing success rate, (c) late decoy-limit penalties, and (d) GPU-hours. Small markers denote attackers, large markers their mean; $\times$: no crossing.}
    \label{fig:rq2}
    \vspace{-7mm}
\end{figure}

\noindent\textbf{Implementation details.} The world model uses a GRU with 1{,}024
deterministic units, a $32{\times}32$ categorical latent, encoders of width 512, and
two-layer MLP heads of width 512.
We optimize the world model with AdamW (learning rate $3{\times}10^{-4}$, weight decay $10^{-4}$, and gradient clipping at 1.0) using batches of 16 windows of 16 steps. We set the dynamics and representation KL weights to 0.5 and 0.1, respectively, use 1 free nat and 1\% uniform mixing, assign a change weight of 5 to node bits, weight the text and frame reconstruction heads by 0.3 and 20, respectively, and define reward-bearing steps by $|r_t| \geq 50$.
\begin{wrapfigure}{r}{0.56\textwidth}
    \centering
    \vspace{-2mm}
    \includegraphics[width=\linewidth]{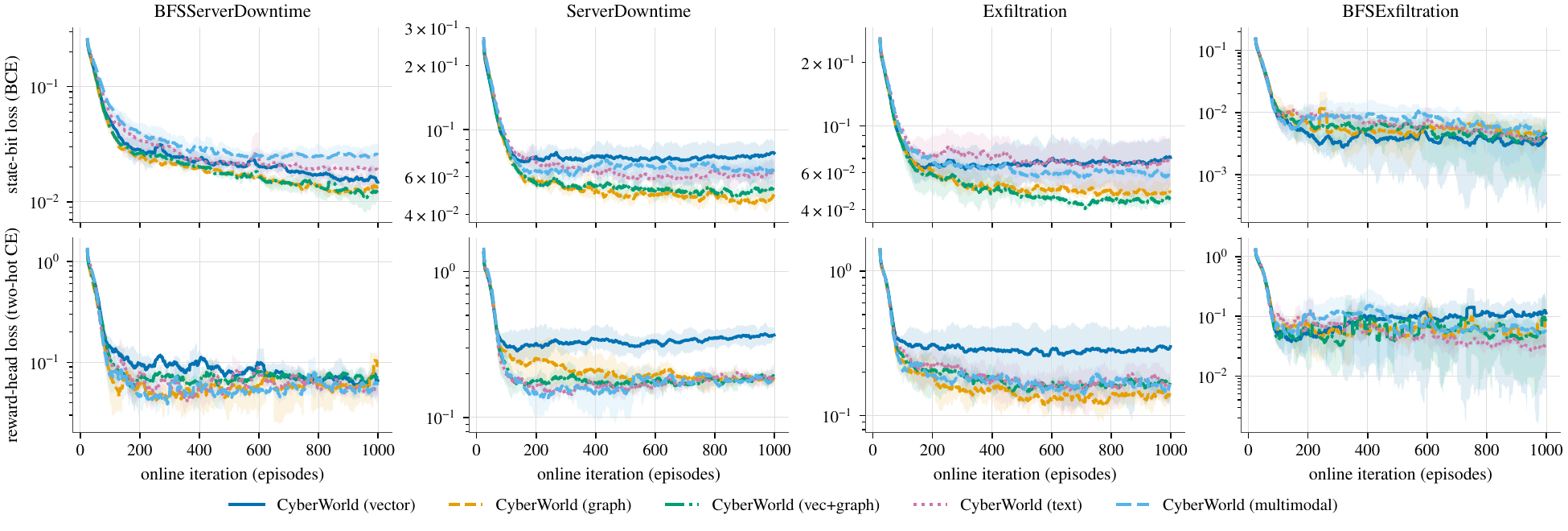} 
    \vspace{-6mm}
    \caption{World-model fidelity across representations at $N{=}15$: replay-batch state-bit loss (top) and reward-head loss (bottom).}
    \label{fig:fidelity}
    \vspace{-6mm}
\end{wrapfigure}
The actor and critic use learning rate $3{\times}10^{-5}$ with a 1{,}000-step warm-up and
cosine decay, imagination horizon 15, $\gamma=0.997$, $\lambda=0.95$, and entropy weight
$3{\times}10^{-4}$. Each run collects a 50-episode random prefill, then one episode per
iteration with 32 world-model and 32 actor--critic updates for 1{,}000 iterations, on a
single NVIDIA A100 shared with other jobs; PPO trains on CPU with 128 parallel
environments.

\vspace{-3mm}
\subsection{Results}
\vspace{-3mm}
\label{sec:exp_results}

\begin{figure}[t]
    \centering
    \includegraphics[width=0.94\linewidth]{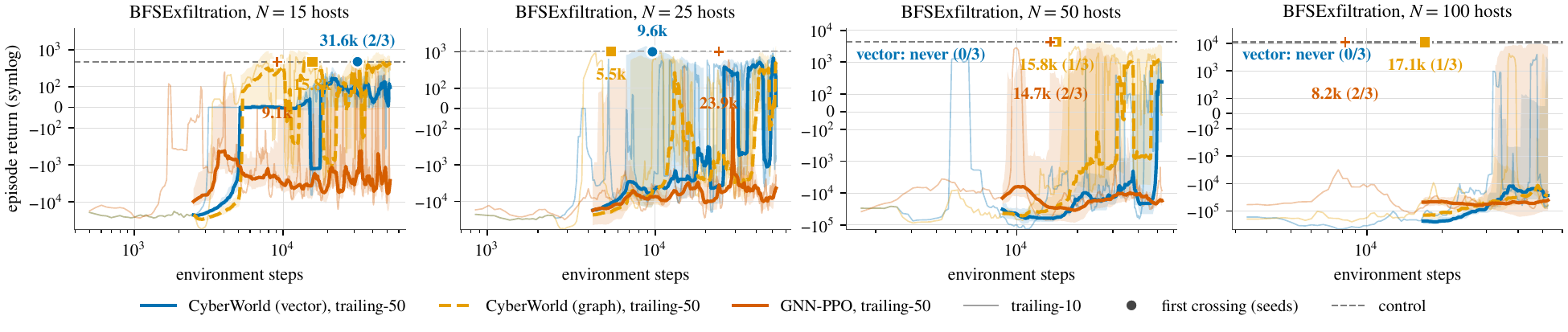}
    \vspace{-3mm}
    \caption{Scaling with network size for the vector, graph, and GNN-PPO arms. Curves show trailing-50 means with min/max bands; thin lines show trailing-10. }
    \label{fig:bfsexfil}
    \vspace{-4mm}
\end{figure}

\noindent\textbf{Sample efficiency (RQ1).}
\autoref{fig:rq1} summarizes the primary sample-efficiency result for the vector and graph arms. On three of the four scoreable attackers, both arms reach the control after $3.6$k--$5.5$k environment steps, whereas PPO requires $2.3$M--$3.1$M steps, two to three orders of magnitude more; on BFSExfiltration, where no PPO seed crosses within its $3.2$M-step budget, the graph arm crosses on every seed at $15{,}750$ steps and the vector arm on two of three seeds at $31{,}650$, while the text arm, not shown, crosses on every seed at $7{,}850$ (exact values in
\autoref{tab:modality}; pooled profile in \autoref{fig:profile}).
GNN-PPO, which shares the graph arm's observation and encoder, exhibits brief early crossings under the trailing-10 criterion, but these are not sustained in the longer-term trend and are followed by severe return degradation under decoy-limit penalties (\autoref{fig:rq2}b,c).
In contrast, the world-model variants maintain
ServerDowntime performance with substantially fewer such penalties. These results
indicate two distinct advantages: relative to PPO, world-model learning reduces the
required environment interaction, while relative to a model-free learner with the same
encoder, it yields greater stability after first reaching the control. These results
concern environment interactions rather than computational cost, which is analyzed
separately below. \autoref{app:smoothing} shows that the main sample-efficiency conclusion is robust to the smoothing convention, while the BFSExfiltration vector crossing is sensitive to this choice.
\vspace{-1mm}
\noindent\textbf{Representation comparison (RQ2).}
On crossing speed, the three numeric arms are indistinguishable within seed variation
on three of the four attackers (\autoref{fig:rq2}(a)), despite clear differences in
world-model fidelity. The graph arm achieves the lowest state-bit and reward-head losses
on the attackers that separate the representations, whereas elevated vector reward loss
coincides with higher decoy-limit penalties (\autoref{fig:rq2}(c), \autoref{fig:fidelity}).
Richer representations provide selective benefits: the text arm crosses on every scoreable
attacker, on all but one of its seeds, and is fastest on BFSExfiltration, whereas the
multimodal arm crosses on every seed of the two non-BFS attackers but on a single seed of
each BFS attacker. 
BFSExfiltration provides the clearest structural distinction among the three numeric representations: because its target order follows the network topology, only the graph arm crosses on every seed at $N{=}15$ (\autoref{fig:bfsexfil}).
% BFSExfiltration provides the clearest structural distinction: because its target order follows the network topology, only the graph arm crosses on every seed at $N{=}15$ (\autoref{fig:bfsexfil}).
Zero-control attackers are omitted (\autoref{app:rq2}).

\begin{wrapfigure}{r}{0.56\textwidth}
    \vspace{-3mm}
    \centering
    \includegraphics[width=\linewidth]{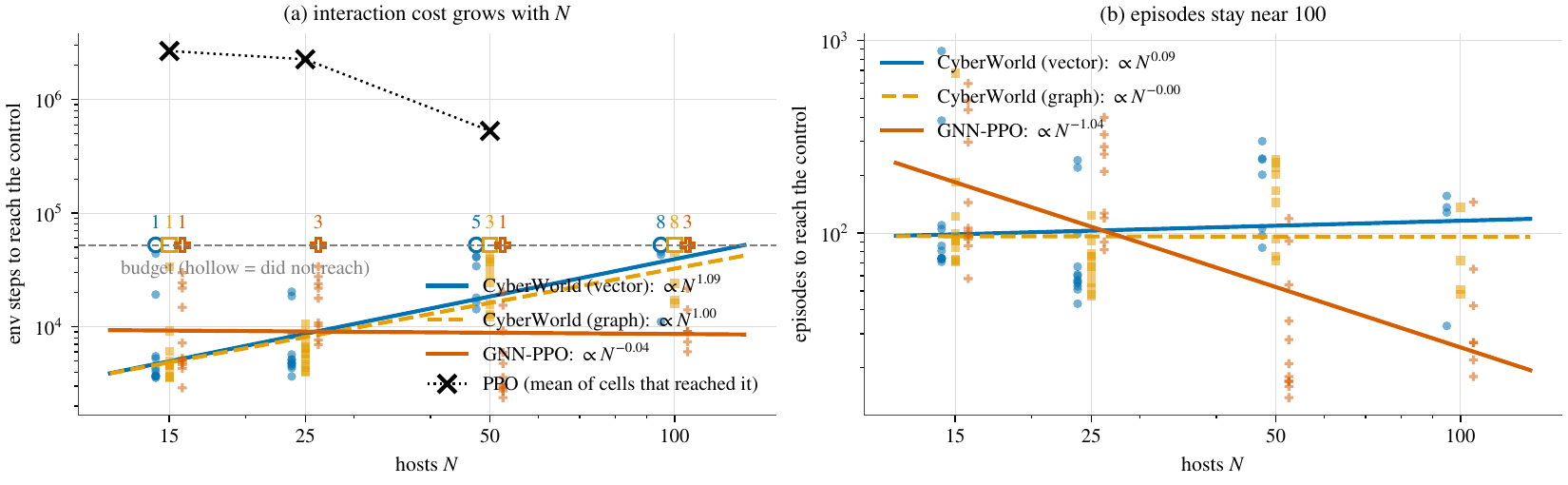}
    \vspace{-4mm}
    \caption{(a) Environment steps and (b) episodes to the control vs.\ $N$ for vector, graph and
    GNN-PPO: small markers are runs that crossed, lines are power-law fits over them,
    hollow markers at the budget line count runs that did not cross, and $\times$ is
    PPO's mean where it crossed.}
    \label{fig:rq3}
    \vspace{-4mm}
\end{wrapfigure}

\noindent\textbf{Scalability (RQ3).}
Among successful runs, the number of episodes required to reach the control remains
approximately constant as the network grows, while environment steps increase with
episode length (\autoref{fig:rq3}). Scalability becomes representation-dependent under
BFSExfiltration: vector and graph both reach the control at $N\leq25$, whereas among the world-model arms only the
graph arm does so at $N{=}50$ and $N{=}100$ (\autoref{fig:bfsexfil}). At $N{=}100$, the
fixed interaction budget covers relatively few episodes, making crossing frequency
increasingly budget-limited. GNN-PPO's trailing-10 return also crosses at these scales, but its trailing-50 trend does not remain above the control once decoy-limit penalties accumulate. Thus, simple
representations suffice for most attackers, while explicit graph structure becomes
increasingly important when attack progression depends on network topology.
See Appendix~\ref{app:scaling}.
% \noindent\textbf{Scalability (RQ3).}
% {\autoref{fig:rq3} follows the numeric arms to larger networks (values in
% \autoref{tab:scale}, curves in \autoref{fig:scale_curves}). Over the runs that crossed,
% environment steps to the control grow in proportion to the episode length while the
% number of episodes stays approximately constant: larger networks lengthen the episodes
% without making adaptation harder. Whether a run crosses at all depends on the
% representation when the attacker follows the topology. \autoref{fig:bfsexfil} shows the
% progression: both arms learn BFSExfiltration at $N \le 25$, and the graph arm alone still
% reaches the control at $N{=}50$ and $N{=}100$. At $N{=}100$ the fixed budget covers few
% episodes and crossings become rare for every arm, so that column measures the budget
% rather than the method. GNN-PPO crosses early at large $N$ but only transiently, its
% late return falling far below the control under decoy-limit penalties. Simple
% representations therefore suffice for most of the dynamics studied here, explicit graph
% structure becomes necessary when the task dynamics depend on the network structure, and
% that dependence sharpens with scale.
\vspace{-3mm}
\paragraph{Discussion and Limitations.}
CyberWorld reaches effective defense with two to three orders of magnitude fewer
environment interactions than PPO, exhibits greater post-crossing stability than a
representation-matched model-free baseline, and, among successful runs, requires an approximately constant number of episodes to reach the control as the network grows from 15 to 100 hosts. 
Among the evaluated representations,
the host graph provides the most consistent performance under topology-dependent attacks.
These results are subject to three main limitations. First, interaction efficiency does not
imply equivalent wall-clock efficiency, particularly for text and multimodal variants that
incur substantially higher inference cost. Second, our evaluation considers an RSSM-based
latent dynamics model and does not compare alternative world-model backbones such as
transformers or diffusion models. Third, all experiments are conducted in CyberWheel with
scripted adversaries, so broader validation across environments and adversary models
remains necessary.

%% file: tables_w_gnn_ppo/T2_scale.tex
\begin{table}[t]
\centering
\caption{Environment steps until the trailing-10 return first exceeds the control.
All methods start from scratch. CyberWorld and GNN-PPO share the same interaction
budget at each $N$; PPO uses its original, longer training horizon. Mean $\pm$ sd
over 3 seeds; $(k/n)$: seeds that crossed. Bold: fastest of GNN-PPO, vector, and
graph with all seeds crossed. PPO: 3 seeds at $N\leq50$, none at
$N=100$; GNN-PPO: 3 seeds, two BFSServerDowntime seeds unfinished at $N=100$.}
\vspace{-2mm}
\label{tab:scale}

\begin{adjustbox}{
    max width=0.94\textwidth,
    max totalheight=0.82\textheight,
    keepaspectratio,
    center
}

\small
\setlength{\tabcolsep}{4pt}
\begin{tabular}{llrrrr}
\toprule
attacker & $N$ & PPO & GNN-PPO & ours: vector & ours: graph \\
\midrule
\multirow{4}{*}{BFSServerDowntime}
 & 15 & 2{,}310{,}400 $\pm$ 640{,}000 (2/3) & \textbf{4{,}017 $\pm$ 821} & 4{,}933 $\pm$ 633 & 5{,}525 $\pm$ 575 (2/3) \\
 & 25 & 2{,}549{,}547 $\pm$ 410{,}310 & 12{,}368 $\pm$ 5{,}398 (2/3) & 10{,}257 $\pm$ 7{,}123 & \textbf{5{,}100 $\pm$ 524} \\
 & 50 & NEVER (0/3) & \textbf{2{,}777 $\pm$ 289} & 37{,}570 $\pm$ 3{,}400 (2/3) & 32{,}725 $\pm$ 8{,}245 (2/3) \\
 & 100 & -- & NEVER (0/1) & NEVER (0/3) & NEVER (0/3) \\
\midrule
\multirow{4}{*}{ServerDowntime}
 & 15 & 3{,}078{,}400 $\pm$ 0 (1/3) & 25{,}875 $\pm$ 4{,}025 (2/3) & 3{,}633 $\pm$ 62 & \textbf{3{,}600 $\pm$ 41} \\
 & 25 & 3{,}274{,}880 $\pm$ 1{,}579{,}162 & 21{,}845 $\pm$ 0 (1/3) & \textbf{5{,}922 $\pm$ 1{,}769} & 7{,}027 $\pm$ 2{,}668 \\
 & 50 & 456{,}960 $\pm$ 0 & \textbf{3{,}457 $\pm$ 924} & 24{,}593 $\pm$ 12{,}038 & 29{,}580 $\pm$ 12{,}300 \\
 & 100 & -- & 35{,}175 $\pm$ 13{,}400 (2/3) & 47{,}570 $\pm$ 4{,}690 (2/3) & 16{,}080 $\pm$ 0 (1/3) \\
\midrule
\multirow{4}{*}{Exfiltration}
 & 15 & 2{,}630{,}400 $\pm$ 448{,}000 (2/3) & 11{,}367 $\pm$ 9{,}182 & \textbf{3{,}900 $\pm$ 283} & 4{,}600 $\pm$ 41 \\
 & 25 & 373{,}547 $\pm$ 102{,}578 & 14{,}110 $\pm$ 7{,}030 & \textbf{4{,}222 $\pm$ 424} & 6{,}290 $\pm$ 367 \\
 & 50 & 602{,}027 $\pm$ 205{,}155 & \textbf{8{,}330 $\pm$ 5{,}141} & 33{,}745 $\pm$ 17{,}255 (2/3) & 25{,}103 $\pm$ 8{,}951 \\
 & 100 & -- & \textbf{9{,}715 $\pm$ 3{,}316} & 28{,}308 $\pm$ 17{,}252 (2/3) & 34{,}840 $\pm$ 10{,}720 (2/3) \\
\midrule
\multirow{4}{*}{BFSExfiltration}
 & 15 & NEVER (0/3) & \textbf{9{,}117 $\pm$ 4{,}128} & 31{,}650 $\pm$ 12{,}450 (2/3) & 15{,}750 $\pm$ 12{,}960 \\
 & 25 & 2{,}839{,}680 $\pm$ 615{,}466 & 23{,}913 $\pm$ 10{,}031 & 9{,}633 $\pm$ 6{,}351 & \textbf{5{,}468 $\pm$ 1{,}627} \\
 & 50 & NEVER (0/3) & 14{,}705 $\pm$ 5{,}525 (2/3) & NEVER (0/3) & 15{,}810 $\pm$ 0 (1/3) \\
 & 100 & -- & 8{,}208 $\pm$ 838 (2/3) & NEVER (0/3) & 17{,}085 $\pm$ 0 (1/3) \\
\bottomrule
\end{tabular}
\end{adjustbox}
\vspace{-7mm}
\end{table}

%% file: sections/conclusion.tex
\vspace{-3mm}
\section{Conclusion}
\vspace{-3mm}

We introduced \textbf{CyberWorld}, a world-model-based framework for autonomous cyber
defense that learns predictive latent dynamics of the defended network and optimizes
defensive policies through imagined trajectories. Our results show that explicit dynamics
modeling can substantially reduce environment interaction while maintaining effective
defensive behavior across adversary strategies and network scales, and that representation
choice materially affects this capability when attack progression depends on network
topology. Future work should extend CyberWorld to alternative latent dynamics architectures,
more efficient semantic representations, additional cyber-defense environments, emulated
networks, and learned adversaries, with the broader goal of developing transferable cyber
world models that capture reusable network dynamics across heterogeneous operational
settings.

%% file: appendices/appendix_B_theory.tex
% ============================================================
% Theoretical analysis of mechanism invariance for CyberWorld.
% Drop-in: insert the main-text part after Section 3.1 (after Eq. 1
% and the b_t discussion), and the proofs as an appendix section.
% Requires: amsthm-style environments. Add to preamble if missing:
%   \newtheorem{assumption}{Assumption}
%   \newtheorem{lemma}{Lemma}
%   \newtheorem{proposition}{Proposition}
%   \newtheorem{theorem}{Theorem}
%   \newtheorem{corollary}{Corollary}
%   \newtheorem{remark}{Remark}
% ============================================================

% ------------------------------------------------------------
% MAIN TEXT
% ------------------------------------------------------------
\section{Theoretical Analysis of the Problem Formulation}
\label{app:theory}

\subsection{Mechanism Invariance and Transfer Across Adversary Strategies}

Equation~(1) decomposes the defender-facing transition law into a
strategy-dependent adversary policy $\pi^R_c$ and a shared mechanism kernel
$\mathcal{T}_{\mathrm{mech}}$. We formalize three consequences of this
factorization. First, changes in the effective transition law are bounded by
changes in the adversary strategy. Second, under partial observability, the
filtering posterior provides a sufficient statistic for defender control,
motivating the use of a recurrent latent state. Third, model error under a
new adversary can be decomposed into mechanism-model error and
strategy-estimation error. Proofs are provided in
\autoref{app:proofs}.

\paragraph{Setup.}
Consider a family of POMDPs $\{\mathcal{M}_c\}_{c \in \mathcal{C}}$,
\[
\mathcal{M}_c =
(\mathcal{S}, \mathcal{A}^B, \mathcal{O},
\mathcal{T}_c, \Omega, R, \gamma),
\]
that share all components except the effective transition law
$\mathcal{T}_c$, which factors according to Eq.~(1). Rewards are bounded as
$|R(s,a^B)| \le R_{\max}$, with $R_{\max}=5{,}000$ in CyberWheel, and
$\gamma \in [0,1)$. We write $\mathrm{TV}(\mu,\nu)$ for total-variation
distance and $V^\pi_{\mathcal{M}}$ for the value of policy $\pi$ in model
$\mathcal{M}$, with the initial-state distribution suppressed from the
notation.

\begin{assumption}[Closed information state]
\label{ass:closed}
The global state $s \in \mathcal{S}$ includes the adversary's internal
belief $b^R$ (``latent adversarial intent'' in Section~3.1), such that
$\pi^R_c(a^R \mid b^R)$ is a measurable function of $s$ and Eq.~(1)
defines a Markov kernel on $\mathcal{S}$ for every $c$.
\end{assumption}

\begin{assumption}[Shared mechanisms]
\label{ass:mech}
$\mathcal{T}_{\mathrm{mech}}$, $\Omega$, and $R$ are identical across
$c \in \mathcal{C}$; adversary strategies differ only through
$\pi^R_c$.
\end{assumption}

Assumption~\ref{ass:mech} formalizes the intended separation between
strategy and environment mechanisms: protocol semantics, action effects,
exploit preconditions, and decoy behavior are assumed to remain unchanged
when the adversary strategy changes. Assumption~\ref{ass:closed} is a state
augmentation used to express the coupled defender--adversary process as a
Markov process.

\paragraph{Sensitivity of the effective dynamics to strategy shift.}

\begin{lemma}[Strategy-shift bound]
\label{lem:shift}
Under Assumptions~\ref{ass:closed}--\ref{ass:mech}, for every
$s \in \mathcal{S}$, $a^B \in \mathcal{A}^B$, and
$c,c' \in \mathcal{C}$,
\[
\mathrm{TV}\!\left(
\mathcal{T}_c(\cdot \mid s,a^B),
\mathcal{T}_{c'}(\cdot \mid s,a^B)
\right)
\le
\mathrm{TV}\!\left(
\pi^R_c(\cdot \mid b^R(s)),
\pi^R_{c'}(\cdot \mid b^R(s))
\right).
\]
\end{lemma}

Lemma~\ref{lem:shift} shows that, under the shared-mechanism
factorization, a change in adversary strategy cannot induce a larger
total-variation change in the effective transition kernel than the
corresponding change in the adversary action distribution. Thus, strategy
variation affects the defender-facing dynamics through a constrained
component of the transition model, while
$\mathcal{T}_{\mathrm{mech}}$ remains shared by assumption.

\paragraph{Belief-state sufficiency under partial observability.}

\begin{proposition}[Belief-state sufficiency]
\label{prop:belief}
For every $c$, the filtering posterior
\[
b_t \doteq p(s_t \mid o_{\le t},a^B_{<t})
\]
is a sufficient statistic of the interaction history. In particular, there
exists an optimal defender policy of the form
$\pi^\star(a^B_t \mid b_t)$, and the belief state evolves according to a
Bayes operator
\[
b_{t+1}=U(b_t,a^B_t,o_{t+1}),
\]
whose dependence on $c$ enters through $\mathcal{T}_c$, and hence through
$\pi^R_c$ under Assumption~\ref{ass:mech}
\citep{aastrom1965optimal,kaelbling1998planning}.
\end{proposition}

Proposition~\ref{prop:belief} provides a control-theoretic interpretation
of the recurrent latent state $s_t=[h_t;z_t]$. An exact belief state is
generally unavailable, and CyberWorld does not require the learned latent
state to recover $b_t$ exactly. Rather, the RSSM can be viewed as learning
a history-dependent predictive statistic from which the actor and value
functions are computed. Its prior
$p_\phi(z_t\mid h_t)$ and posterior
$q_\phi(z_t\mid h_t,e_t)$ are respectively analogous to the prediction and
observation-update stages of Bayesian filtering. This interpretation
motivates RQ2: different observation representations may preserve different
amounts of information relevant to constructing such a predictive latent
state.

\paragraph{Error decomposition under adversary shift.}

The following decomposition is an analytical abstraction of the transition
factorization in Equation~(1); CyberWorld does not explicitly parameterize
$\widehat{\mathcal{T}}_{\mathrm{mech}}$ and $\widehat{\pi}^R_c$ as separate
learned modules. To characterize the potential benefit of shared mechanisms,
suppose an idealized factored predictive model has obtained a mechanism model
$\widehat{\mathcal{T}}_{\mathrm{mech}}$ from data generated under one or
more adversary strategies and, for a possibly new adversary $c'$, an
estimate $\widehat{\pi}^R_{c'}$ of its strategy. Define the corresponding
imagined dynamics as
\[
\widehat{\mathcal{T}}_{c'}(s' \mid s,a^B)
=
\sum_{a^R}
\widehat{\mathcal{T}}_{\mathrm{mech}}
(s' \mid s,a^B,a^R)
\widehat{\pi}^R_{c'}(a^R \mid b^R(s)),
\]
with error radii
\[
\varepsilon_{\mathrm{mech}}
\doteq
\sup_{s,a^B,a^R}
\mathrm{TV}\!\left(
\mathcal{T}_{\mathrm{mech}}(\cdot\mid s,a^B,a^R),
\widehat{\mathcal{T}}_{\mathrm{mech}}(\cdot\mid s,a^B,a^R)
\right),
\]
and
\[
\varepsilon_{\mathrm{strat}}(c')
\doteq
\sup_s
\mathrm{TV}\!\left(
\pi^R_{c'}(\cdot\mid b^R(s)),
\widehat{\pi}^R_{c'}(\cdot\mid b^R(s))
\right).
\]

\begin{theorem}[Decomposed simulation bound]
\label{thm:transfer}
Under Assumptions~\ref{ass:closed}--\ref{ass:mech}, for every stationary
defender policy $\pi$ and every $c' \in \mathcal{C}$,
\[
\left|
V^\pi_{\mathcal{M}_{c'}}
-
V^\pi_{\widehat{\mathcal{M}}_{c'}}
\right|
\le
\frac{2\gamma R_{\max}}{(1-\gamma)^2}
\left(
\varepsilon_{\mathrm{mech}}
+
\varepsilon_{\mathrm{strat}}(c')
\right).
\]
Consequently, if
$\widehat{\pi}\in
\arg\max_\pi V^\pi_{\widehat{\mathcal{M}}_{c'}}$,
then
\[
V^{\pi^\star}_{\mathcal{M}_{c'}}
-
V^{\widehat{\pi}}_{\mathcal{M}_{c'}}
\le
\frac{4\gamma R_{\max}}{(1-\gamma)^2}
\left(
\varepsilon_{\mathrm{mech}}
+
\varepsilon_{\mathrm{strat}}(c')
\right).
\]
\end{theorem}

Theorem~\ref{thm:transfer} separates two sources of error. The term
$\varepsilon_{\mathrm{mech}}$ measures error in the component assumed to
be shared across adversary strategies, whereas
$\varepsilon_{\mathrm{strat}}(c')$ measures error associated with the
particular adversary $c'$. Consequently, improvements to
$\widehat{\mathcal{T}}_{\mathrm{mech}}$ can, under
Assumption~\ref{ass:mech}, benefit predictions under multiple adversary
strategies. Adaptation to $c'$ nevertheless remains dependent on how
accurately its strategy can be represented or inferred.

\begin{corollary}[Residual error under a previously learned mechanism]
\label{cor:adapt}
Fix a target suboptimality $\epsilon$ and suppose
\[
\varepsilon_{\mathrm{mech}}
\le
\frac{(1-\gamma)^2}{8\gamma R_{\max}}\epsilon.
\]
Then the bound in Theorem~\ref{thm:transfer} guarantees suboptimality at
most $\epsilon$ whenever
\[
\varepsilon_{\mathrm{strat}}(c')
\le
\frac{(1-\gamma)^2}{8\gamma R_{\max}}\epsilon.
\]
Thus, once the mechanism error is below the stated threshold, the bound
requires no further reduction in $\varepsilon_{\mathrm{mech}}$ for this
target accuracy; the remaining contribution is
$\varepsilon_{\mathrm{strat}}(c')$.
\end{corollary}

Notably, the bound contains no explicit dependence on the number of hosts
$N$. This does not imply that adaptation is independent of network size:
larger networks may make state estimation, belief tracking, function
approximation, or data collection more difficult. Rather, under the stated
factorization, network size does not introduce an additional term into the
error decomposition itself. The approximately stable episodes-to-control
observed from $15$ to $100$ hosts in RQ3 are empirically consistent with
this interpretation, while not constituting a direct consequence of the
bound.

\begin{remark}[Sensitivity of a fixed policy to strategy shift]
\label{rem:modelfree}
For a fixed policy $\pi_c$, the same simulation argument gives
\[
\left|
V^{\pi_c}_{\mathcal{M}_c}
-
V^{\pi_c}_{\mathcal{M}_{c'}}
\right|
\le
\frac{2\gamma R_{\max}}{(1-\gamma)^2}
\sup_s
\mathrm{TV}\!\left(
\pi^R_c(\cdot\mid b^R(s)),
\pi^R_{c'}(\cdot\mid b^R(s))
\right).
\]
This bound quantifies the sensitivity of a fixed policy to an adversary
    shift. It does not, by itself, establish the interaction complexity
    required to adapt a model-free agent to $c'$. The distinction relevant to
    Theorem~\ref{thm:transfer} is therefore analytical: under the factorization
    of Equation~(1), mechanism information constitutes a shared component that
    can in principle remain useful when the strategy-dependent component changes.
% This bound quantifies the sensitivity of a fixed policy to an adversary
% shift. It does not, by itself, establish the interaction complexity
% required to adapt a model-free agent to $c'$. The distinction relevant to
% Theorem~\ref{thm:transfer} is instead representational: the factored model
% retains an explicit shared component
% $\widehat{\mathcal{T}}_{\mathrm{mech}}$, allowing previously learned
% mechanism information to remain available when the strategy-dependent
% component changes.
\end{remark}

% ------------------------------------------------------------
% APPENDIX: proofs
% ------------------------------------------------------------
\section{Proofs for Section~\ref{app:theory}}
\label{app:proofs}

\subsection{Proof of Lemma~\ref{lem:shift}}

Fix $s,a^B$ and abbreviate $b=b^R(s)$ and
\[
K_{a^R}(\cdot)
=
\mathcal{T}_{\mathrm{mech}}
(\cdot\mid s,a^B,a^R).
\]
By Eq.~(1), for any measurable $E\subseteq\mathcal{S}$,
\[
\mathcal{T}_c(E\mid s,a^B)
-
\mathcal{T}_{c'}(E\mid s,a^B)
=
\sum_{a^R}
K_{a^R}(E)
\left[
\pi^R_c(a^R\mid b)
-
\pi^R_{c'}(a^R\mid b)
\right].
\]
Since $0\le K_{a^R}(E)\le1$, taking the supremum over $E$ and using
\[
\mathrm{TV}(\mu,\nu)
=
\sup_E |\mu(E)-\nu(E)|
\]
together with
\[
\sup_E
\left|
\sum_i w_i K_i(E)
\right|
\le
\frac{1}{2}\sum_i |w_i|
\]
for signed weights satisfying $\sum_iw_i=0$ yields
\[
\mathrm{TV}
\!\left(
\mathcal{T}_c(\cdot\mid s,a^B),
\mathcal{T}_{c'}(\cdot\mid s,a^B)
\right)
\le
\frac{1}{2}
\sum_{a^R}
\left|
\pi^R_c(a^R\mid b)
-
\pi^R_{c'}(a^R\mid b)
\right|.
\]
The right-hand side is
$\mathrm{TV}(\pi^R_c(\cdot\mid b),
\pi^R_{c'}(\cdot\mid b))$, which proves the result.
\qed

\subsection{Proof of Proposition~\ref{prop:belief}}

Sufficiency of the filtering posterior and the existence of an optimal
belief-measurable policy are standard results for POMDPs. The belief update is
\[
b_{t+1}(s')
\propto
\Omega(o_{t+1}\mid s')
\sum_s
\mathcal{T}_c(s'\mid s,a^B_t)b_t(s).
\]
Under Assumption~\ref{ass:mech}, the dependence on $c$ enters only through
$\mathcal{T}_c$, which by Eq.~(1) depends on the adversary strategy
$\pi^R_c$. Hence $b_t$ is a sufficient statistic for optimal control and
evolves according to a Bayes operator of the stated form.
\qed

The proposition concerns the exact filtering posterior of the POMDP. The
RSSM interpretation in \autoref{app:theory} is approximate: CyberWorld
does not assume that $s_t=[h_t;z_t]$ recovers $b_t$, nor do the RSSM losses
establish such recovery. Rather,
$p_\phi(z_t\mid h_t)$ and
$q_\phi(z_t\mid h_t,e_t)$ play roles analogous to predictive and
observation-conditioned latent distributions, respectively, providing a
learned recurrent representation on which the policy is conditioned.

\subsection{Proof of Theorem~\ref{thm:transfer}}

\paragraph{Step 1: One-step model error.}
Introduce the hybrid kernel
\[
\widetilde{\mathcal{T}}(s'\mid s,a^B)
=
\sum_{a^R}
\widehat{\mathcal{T}}_{\mathrm{mech}}
(s'\mid s,a^B,a^R)
\pi^R_{c'}(a^R\mid b^R(s)),
\]
which combines the learned mechanism model with the true adversary
strategy. By the triangle inequality,
\[
\mathrm{TV}
\!\left(
\mathcal{T}_{c'},
\widehat{\mathcal{T}}_{c'}
\right)
\le
\mathrm{TV}
\!\left(
\mathcal{T}_{c'},
\widetilde{\mathcal{T}}
\right)
+
\mathrm{TV}
\!\left(
\widetilde{\mathcal{T}},
\widehat{\mathcal{T}}_{c'}
\right).
\]
The first term is bounded by
$\varepsilon_{\mathrm{mech}}$ because it averages mechanism-model
deviations under
$\pi^R_{c'}(\cdot\mid b^R(s))$. The second term is bounded by
$\varepsilon_{\mathrm{strat}}(c')$ by the same mixture-contraction
argument as Lemma~\ref{lem:shift}. Therefore,
\[
\mathrm{TV}
\!\left(
\mathcal{T}_{c'}(\cdot\mid s,a^B),
\widehat{\mathcal{T}}_{c'}(\cdot\mid s,a^B)
\right)
\le
\varepsilon_{\mathrm{mech}}
+
\varepsilon_{\mathrm{strat}}(c').
\]

\paragraph{Step 2: Simulation bound.}
Let
\[
\varepsilon
\doteq
\varepsilon_{\mathrm{mech}}
+
\varepsilon_{\mathrm{strat}}(c').
\]
For any stationary policy $\pi$,
$\|V^\pi\|_\infty\le R_{\max}/(1-\gamma)$. A standard simulation-lemma
argument \citep{kearns2002near} gives
\[
\left|
V^\pi_{\mathcal{M}_{c'}}
-
V^\pi_{\widehat{\mathcal{M}}_{c'}}
\right|
\le
\sum_{t\ge0}
\gamma^{t+1}
\,
2\varepsilon
\,
\frac{R_{\max}}{1-\gamma}.
\]
Summing the geometric series yields
\[
\left|
V^\pi_{\mathcal{M}_{c'}}
-
V^\pi_{\widehat{\mathcal{M}}_{c'}}
\right|
\le
\frac{2\gamma R_{\max}}{(1-\gamma)^2}
\varepsilon.
\]

\paragraph{Step 3: Policy suboptimality.}
Let
\[
\widehat{\pi}
\in
\arg\max_\pi
V^\pi_{\widehat{\mathcal{M}}_{c'}}.
\]
Then
\[
\begin{aligned}
V^{\pi^\star}_{\mathcal{M}_{c'}}
-
V^{\widehat{\pi}}_{\mathcal{M}_{c'}}
&=
\left(
V^{\pi^\star}_{\mathcal{M}_{c'}}
-
V^{\pi^\star}_{\widehat{\mathcal{M}}_{c'}}
\right)
\\
&\quad+
\left(
V^{\pi^\star}_{\widehat{\mathcal{M}}_{c'}}
-
V^{\widehat{\pi}}_{\widehat{\mathcal{M}}_{c'}}
\right)
\\
&\quad+
\left(
V^{\widehat{\pi}}_{\widehat{\mathcal{M}}_{c'}}
-
V^{\widehat{\pi}}_{\mathcal{M}_{c'}}
\right).
\end{aligned}
\]
The middle term is non-positive by optimality of $\widehat{\pi}$ in
$\widehat{\mathcal{M}}_{c'}$, and the first and third terms are each
bounded by Step~2. Hence
\[
V^{\pi^\star}_{\mathcal{M}_{c'}}
-
V^{\widehat{\pi}}_{\mathcal{M}_{c'}}
\le
\frac{4\gamma R_{\max}}{(1-\gamma)^2}
\left(
\varepsilon_{\mathrm{mech}}
+
\varepsilon_{\mathrm{strat}}(c')
\right).
\]
\qed

\subsection{Proof of Remark~\ref{rem:modelfree}}

Fix a policy $\pi_c$ and compare its value under
$\mathcal{M}_c$ and $\mathcal{M}_{c'}$. Applying the simulation argument
from Theorem~\ref{thm:transfer} together with
Lemma~\ref{lem:shift} gives
\[
\left|
V^{\pi_c}_{\mathcal{M}_c}
-
V^{\pi_c}_{\mathcal{M}_{c'}}
\right|
\le
\frac{2\gamma R_{\max}}{(1-\gamma)^2}
\sup_s
\mathrm{TV}\!\left(
\pi^R_c(\cdot\mid b^R(s)),
\pi^R_{c'}(\cdot\mid b^R(s))
\right).
\]
This establishes sensitivity of a fixed policy to a change in adversary
strategy. The inequality does not imply a particular sample complexity for
retraining or adaptation under $c'$. Its role here is to contrast the
fixed-policy sensitivity bound with the explicit mechanism--strategy
decomposition available in Theorem~\ref{thm:transfer}.
\qed

%% file: appendices/appendix_A.tex
% Supplementary experiments.
\section{Crossing Statistics and Performance Profiles at $N{=}15$}
\label{app:rq1}

\begin{figure}[h]
    \centering
    \includegraphics[width=0.62\linewidth]{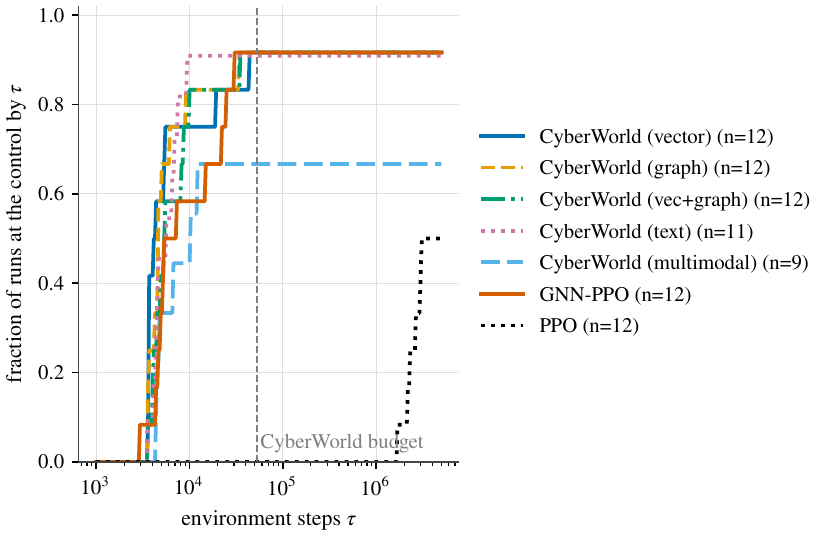}
    \caption{Performance profile at $N{=}15$: fraction of (attacker, seed) runs that
    have reached the control by environment step $\tau$, pooled over the four scoreable
    attackers. Runs that never cross are censored at their budget. Dashed vertical
    line: the $52.5$k-step budget of the world-model runs.}
    \label{fig:profile}
\end{figure}

This appendix provides the measurements behind RQ1 in Section~\ref{sec:exp_results}, in particular the two-to-three-orders-of-magnitude gap to PPO and the first-crossing comparison with GNN-PPO. \autoref{fig:rq1} reports, for each attacker, the environment step at which each method first exceeds the strategy-agnostic control. \autoref{tab:modality} lists the exact values with seed counts, and \autoref{fig:profile} aggregates them across attackers as a performance profile: for each method, the fraction of all (attacker, seed) runs that have reached the control by environment step $\tau$. A profile that rises earlier and higher denotes a more sample-efficient defender, and a run that never crosses remains censored at its budget.

\input{tables_w_gnn_ppo/T1_modality}

The three numeric representations reach the control in 92\% of their runs
within the $52.5$k-step budget, and the text representation does so in all but one of its runs. In contrast, PPO's profile remains at zero until
approximately $2$M steps and reaches only 5 of 12 runs (42\%) by the end of its $3.2$M-step budget. The multimodal representation plateaus lowest among the world-model arms because only one seed crosses on each of the two BFS attackers. GNN-PPO, the representation-matched baseline trained in the same loop, also reaches 92\%, and its profile rises at similar steps to those of the numeric world-model arms; relative to GNN-PPO, the clearest advantage of the world model therefore lies in post-crossing retention rather than initial crossing speed, as \autoref{tab:retention} shows. PPO is particularly seed-sensitive on ServerDowntime: only one of its three seeds ever exceeds the control, whereas every vector and graph seed does so within $4$k steps.
\section[Sensitivity of the Crossing Criterion to the Smoothing Window]{Sensitivity of the Crossing Criterion to the Smoothing Window}
\label{app:smoothing}

\begin{figure}[h]
    \centering
    \includegraphics[width=\linewidth]{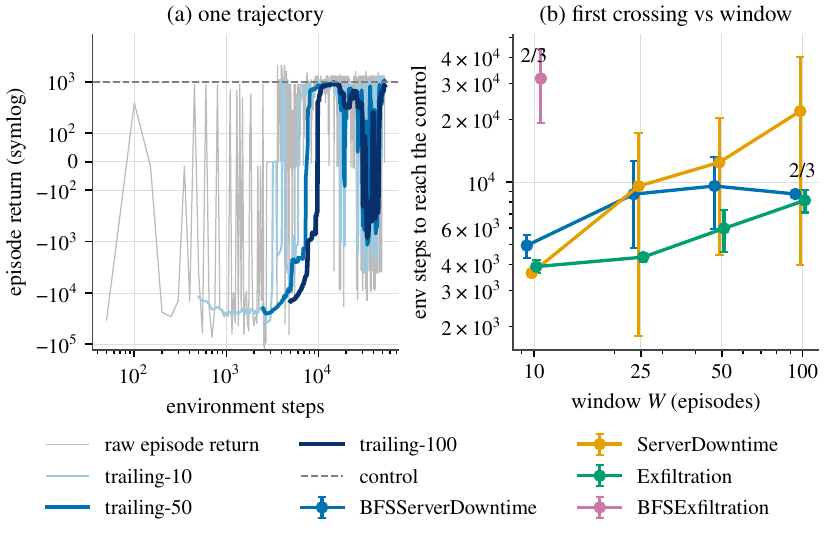}
    \caption{Smoothing-window sensitivity. (a) One trajectory at four temporal
    resolutions. (b) First-crossing step of the vector arm vs.\ the window $W$ used to
    define it (mean $\pm$ sd over 3 seeds; $k/3$ where fewer seeds cross).}
    \label{fig:smoothing}
\end{figure}

This appendix examines how the primary trailing-10 crossing metric changes under alternative smoothing windows and tests whether the main sample-efficiency conclusion depends on this convention. The first crossing is defined on the trailing-10 episode return. Because the environment contains a rare $-5{,}000$ penalty, a short window makes the return curves volatile and could in principle shift the crossing. \autoref{fig:smoothing}a shows one trajectory (vector, BFSServerDowntime, seed 1) as raw returns and at trailing-10, trailing-50, and trailing-100 resolution; \autoref{fig:smoothing}b recomputes the first-crossing step of the vector arm on every attacker under windows $W \in \{10, 25, 50, 100\}$, as mean $\pm$ sd over seeds, annotated with the number of seeds that still cross whenever it falls below three.

In these runs, widening the window generally delays the crossing, but by at most a factor of two on ServerDowntime, Exfiltration, and BFSServerDowntime between $W{=}10$ and $W{=}50$, against a $10^2$--$10^3$ gap to PPO. On BFSExfiltration, where the vector arm settles close to the control after crossing, only $W{=}10$ registers a crossing, and only for two of three seeds; wider windows do not, which indicates that the behavior is acquired but not consistently held. The headline conclusion is therefore independent of the smoothing convention, whereas the BFSExfiltration result depends on it, and the main text reports that cell together with its seed count.

\section{Per-Representation Learning Dynamics and Post-Crossing Retention}
\label{app:rq2}

\begin{figure}[h]
    \centering
    \includegraphics[width=\linewidth]{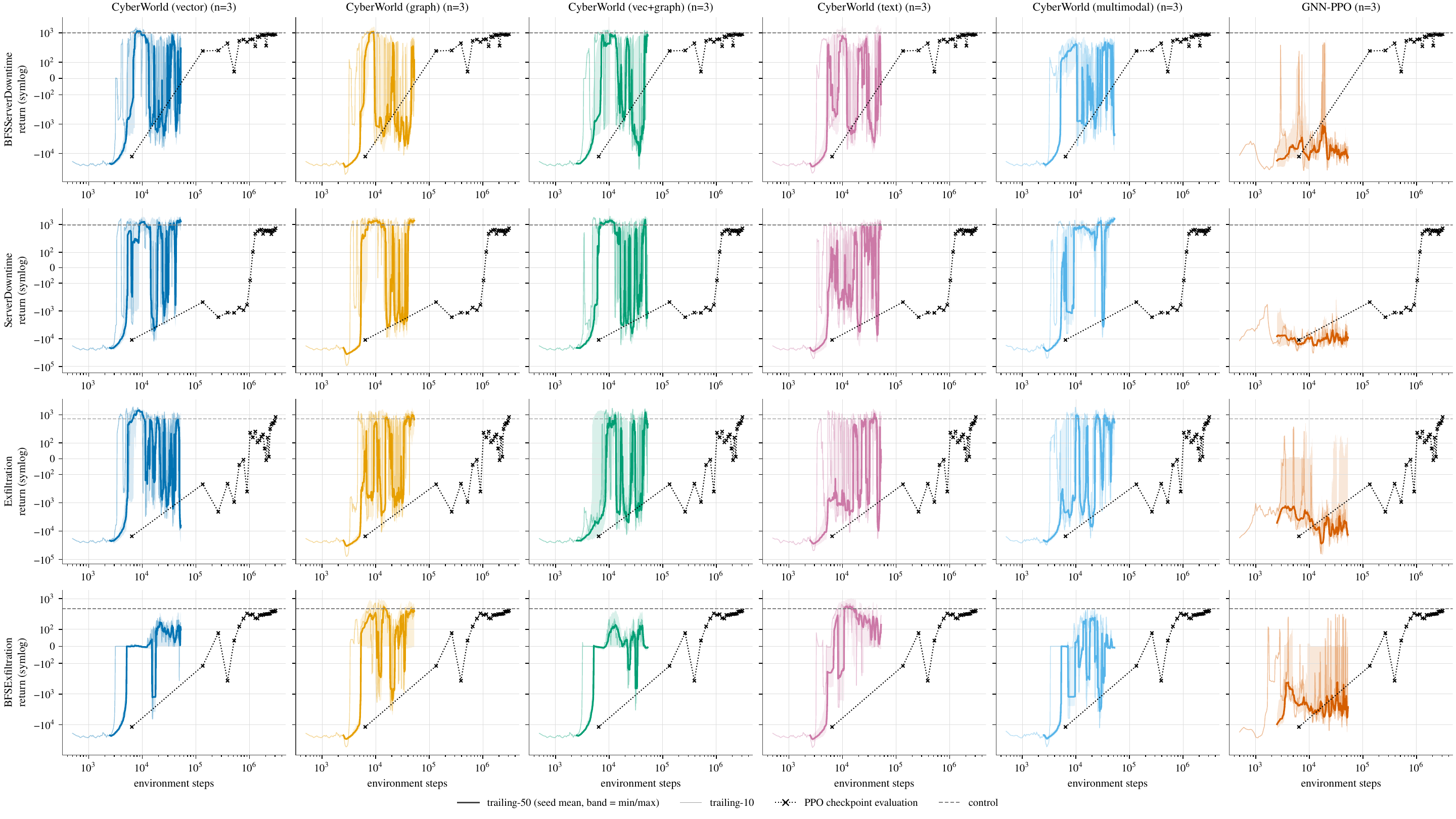}
    \caption{Learning curves of the five representations and GNN-PPO (columns) on the
    four scoreable attackers (rows) at $N{=}15$. Thick: trailing-50 seed mean with min/max
    band; thin: trailing-10; black $\times$: PPO checkpoint evaluations; dashed: control.}
    \label{fig:modality}
\end{figure}

\begin{figure}[h]
    \centering
    \includegraphics[width=0.7\linewidth]{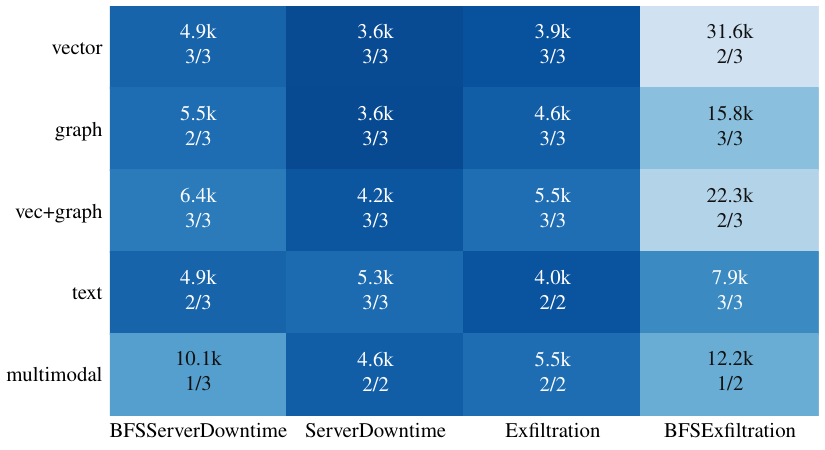}
    \caption{Environment steps to first reach the control per representation and
    attacker at $N{=}15$ (darker = fewer; $k/n$ = seeds that reached it; grey = no seed
    reached it within the $52.5$k-step budget).}
    \label{fig:heatmap}
\end{figure}

This appendix expands the representation comparison of RQ2 in Section~\ref{sec:exp_results}: it provides the complete learning curves summarized in \autoref{fig:rq2} and \autoref{fig:fidelity}, and the post-crossing statistics on which the retention argument of RQ1 (the contrast between the world-model arms and GNN-PPO) rests. \autoref{fig:rq2} summarizes each representation by its crossing step per attacker. This section provides the underlying evidence: \autoref{fig:modality} shows the complete learning curve of every representation on every scoreable attacker, \autoref{fig:heatmap} presents the same crossing steps as a lookup grid, and \autoref{tab:retention} reports what happens after the crossing, namely the share of post-crossing episodes above the control, the median return of the last 200 training episodes, and the rate of decoy-limit penalties in those episodes, for every method including PPO and GNN-PPO.

\label{app:retention}
\input{tables_w_gnn_ppo/A2_retention}

The crossing is the same event for every arm: a transition from the random-policy return of about $-10^4$ to the level of the control within a few hundred episodes. The arms differ in retention, where the comparison with PPO is mixed; \autoref{tab:retention} accordingly carries no bold entries, since the two budgets differ by a factor of $60$. On ServerDowntime the vector and graph arms hold the control in about half of their post-crossing episodes, with a late median above it and fewer than two penalties per 100 episodes, which is on par with PPO after $3$M steps. On Exfiltration the vector and text arms incur the decoy-limit penalty in more than half of their late episodes, whereas the graph and multimodal arms do not. On BFSExfiltration the vec+graph, text, and multimodal arms settle at the do-nothing return of 0 after a brief excursion above it, which corresponds to a collapsed rather than a stable policy. The multimodal arm holds the control with the fewest penalties on ServerDowntime and Exfiltration, at the cost of crossing later or on fewer seeds than the numeric arms. GNN-PPO is the clearest instance of a crossing that does not constitute a defence: its late median return is $-3{,}700$ on both ServerDowntime attackers and 0 on both Exfiltration attackers, with 66 to 237 decoy-limit penalties per 100 late episodes, that is, the policy that crossed continues to deploy beyond the cap.

\section{Scaling Analysis: Crossing Statistics, Learning Curves, and Computational Cost for $N \in \{25, 50, 100\}$}
\label{app:scaling}

\begin{figure}[h]
    \centering
    \includegraphics[width=\linewidth]{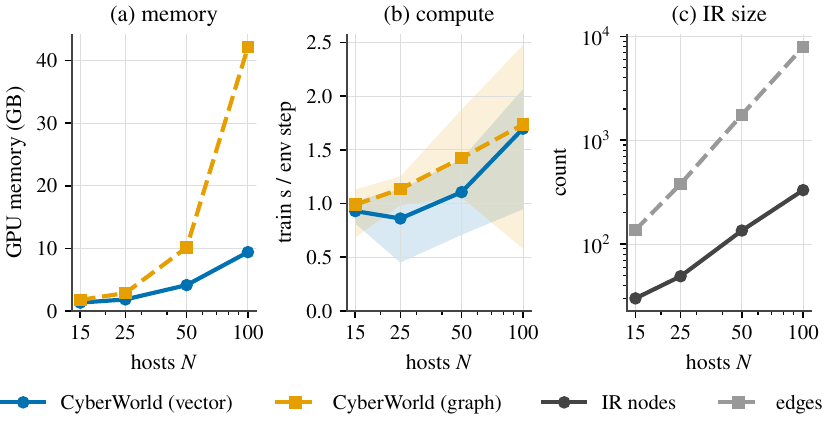}
    \caption{Compute scaling of the two numeric representations. (a) Peak GPU memory
    and (b) training seconds per environment step (median over runs, band = min/max);
    (c) nodes and edges of the intermediate representation vs.\ $N$.}
    \label{fig:compute}
\end{figure}

\begin{figure}
    \centering
    \includegraphics[width=\linewidth]{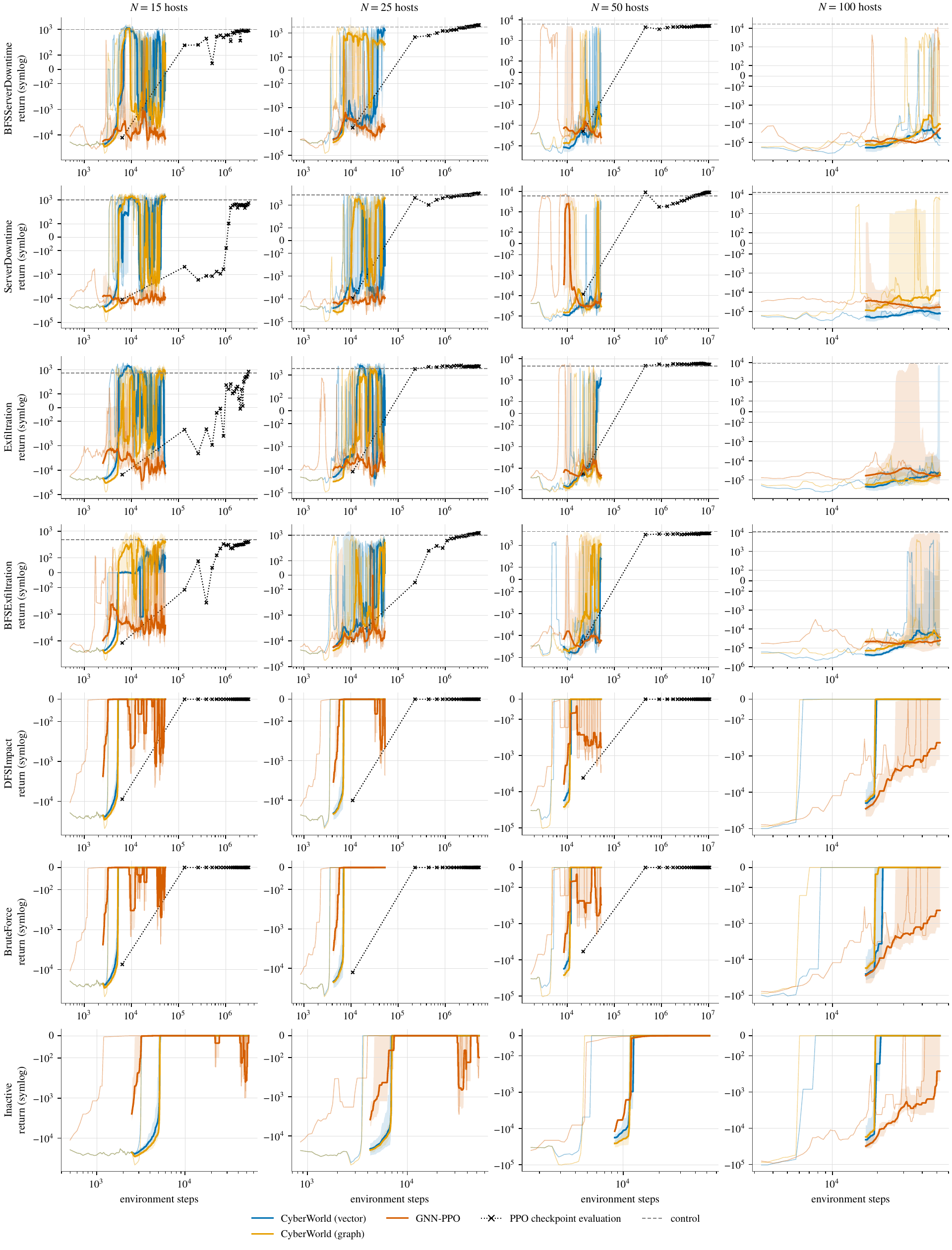}
    \caption{Return vs.\ environment steps for every attacker (rows) and network size
    (columns): CyberWorld vector, graph, and GNN-PPO (trailing-50 seed mean with
    min/max band, thin = trailing-10), PPO checkpoint evaluations, and the control.}
    \label{fig:scale_curves}
\end{figure}

This appendix provides the evidence behind RQ3 in Section~\ref{sec:exp_results}: the exact values summarized in \autoref{fig:rq3} and \autoref{fig:bfsexfil}, every learning curve at $N \in \{25, 50, 100\}$, and the computational cost that the limitations in the Summary paragraph and in Section~\ref{sec:exp_results} refer to. \autoref{fig:rq3} fits the cost of adaptation against $N$ over the runs that crossed. \autoref{tab:scale} in the main text gives the crossing step per attacker and size with seed counts, including PPO at every size at which it was trained. \autoref{fig:compute} reports the cost of each representation as $N$ grows, in terms of peak GPU memory, training seconds per environment step, and the size of the intermediate representation, and \autoref{fig:scale_curves} shows every learning curve for every attacker and size, including the three attackers whose control is 0 and for which no crossing is defined.

At $N{=}25$ the curves resemble those at $N{=}15$ with a longer prefill. At $N{=}50$ the crossing moves to $16$--$38$k steps and the seed spread around the control widens in absolute return, because the control itself is $5$--$6\times$ larger. At $N{=}100$ the $52.5$k-step budget covers only 157 episodes, the curves are still rising when the budget ends, and the wide seed bands indicate that crossing becomes highly seed-dependent; the power-law fit in \autoref{fig:rq3} uses only the runs that did cross and is therefore a statement about the cost of the successful runs rather than a guarantee of success. The graph representation's advantage on the topology-following attacker is obtained at the cost of memory, 43~GB at $N{=}100$ against 10~GB for the vector representation, at similar seconds per environment step. GNN-PPO's curves in \autoref{fig:scale_curves} exhibit the pattern of \autoref{tab:scale} at every size, an early excursion above the control followed by a return that settles far below it, reaching $-9{,}000$ to $-11{,}000$ on the ServerDowntime attackers at $N{=}50$.

%% file: tables_w_gnn_ppo/T1_modality.tex
\begin{table}[h]
\centering
\caption{Environment steps to first reach the \texttt{deploy\_then\_stop} control, 15 hosts,
all methods from scratch. Mean $\pm$ sd over 3 seeds; four text/multimodal cells use 2 seeds
because the third run did not complete; $(k/n)$: seeds that crossed when not all did.
Bold: lowest mean in the column. PPO: raw 30-episode checkpoint evaluations, 3 seeds.
GNN-PPO: the graph arm's observation and GAT encoder trained with PPO instead of the world model,
online training return, 3 seeds. DFSImpact, BruteForce and Inactive are omitted: their control is 0,
so no crossing is defined (all arms end at a final return of 0 there).}
% \caption{Environment steps to first reach the \texttt{deploy\_then\_stop} control, 15 hosts, all methods from scratch. Mean $\pm$ sd over 3 seeds (two for the text and multimodal cells whose third seed did not run); $(k/n)$: seeds that crossed when not all did. Bold: lowest mean in the column. PPO: raw 30-episode checkpoint evaluations, 3 seeds. GNN-PPO: the graph arm's observation and GAT encoder trained with PPO instead of the world model, online training return, 3 seeds. DFSImpact, BruteForce and Inactive are omitted: their control is 0, so no crossing is defined (all arms end at a final return of 0 there).}
\label{tab:modality}

\begin{adjustbox}{
    max width=\textwidth,
    max totalheight=0.82\textheight,
    keepaspectratio,
    center
}

\small
\begin{tabular}{lrrrr}
\toprule
 & BFSServerDowntime & ServerDowntime & Exfiltration & BFSExfiltration \\
\midrule
PPO (from scratch) & 2{,}310{,}400 $\pm$ 640{,}000 (2/3) & 3{,}078{,}400 (1/3) & 2{,}630{,}400 $\pm$ 448{,}000 (2/3) & NEVER (0/3) \\
\midrule
GNN-PPO (from scratch) & \textbf{4{,}017 $\pm$ 821} & 25{,}875 $\pm$ 4{,}025 (2/3) & 11{,}367 $\pm$ 9{,}182 & 9{,}117 $\pm$ 4{,}128 \\
\midrule
ours: vector & 4{,}933 $\pm$ 633 & 3{,}633 $\pm$ 62 & \textbf{3{,}900 $\pm$ 283} & 31{,}650 $\pm$ 12{,}450 (2/3) \\
ours: graph & 5{,}525 $\pm$ 575 (2/3) & \textbf{3{,}600 $\pm$ 41} & 4{,}600 $\pm$ 41 & 15{,}750 $\pm$ 12{,}960 \\
ours: vec+graph & 6{,}433 $\pm$ 1{,}462 & 4{,}183 $\pm$ 554 & 5{,}533 $\pm$ 1{,}897 & 22{,}325 $\pm$ 12{,}525 (2/3) \\
ours: text & 4{,}900 $\pm$ 750 (2/3) & 5{,}283 $\pm$ 1{,}109 & 4{,}000 $\pm$ 400 & \textbf{7{,}850 $\pm$ 1{,}203} \\
ours: multimodal & 10{,}100 $\pm$ 0 (1/3) & 4{,}600 $\pm$ 100 & 5{,}525 $\pm$ 1{,}125 & 12{,}200 $\pm$ 0 (1/2) \\
\bottomrule
\end{tabular}
\end{adjustbox}
\end{table}

%% file: tables_w_gnn_ppo/A2_retention.tex
\begin{table}[h]
\centering
\caption{Retention after the first crossing, 15 hosts. Each cell: share of post-crossing episodes above the control / median return of the last 200 training episodes / decoy-limit penalties per 100 of those episodes. CyberWorld statistics are taken at the end of its $52.5$k-step budget with the stochastic training policy; PPO's are its 30-episode checkpoint evaluations after its own crossing and its last three checkpoints, i.e.\ after up to $3.2$M steps. The two columns therefore measure retention under very different budgets and are not bolded. Seeds are pooled; four text and multimodal cells have two seeds.}
\label{tab:retention}

\begin{adjustbox}{max width=\textwidth,center}
\small
\setlength{\tabcolsep}{4pt}
\begin{tabular}{lrrrr}
\toprule
 & BFSServerDowntime & ServerDowntime & Exfiltration & BFSExfiltration \\
\midrule
PPO & 56\% / 899 / 2.6 & 53\% / 886 / 8.9 & 39\% / 290 / 5.6 & -- / $-1$ / 0.0 \\
GNN-PPO & 23\% / $-3{,}700$ / 237.2 & 28\% / $-3{,}700$ / 220.8 & 12\% / 0 / 167.8 & 17\% / 0 / 66.0 \\
\midrule
CyberWorld (vector) & 33\% / 888 / 23.7 & 51\% / 1{,}192 / 0.8 & 30\% / 285 / 55.0 & 21\% / 0 / 0.5 \\
CyberWorld (graph) & 30\% / 388 / 14.7 & 49\% / 1{,}181 / 1.7 & 32\% / 190 / 2.7 & 37\% / $-4$ / 0.0 \\
CyberWorld (vec+graph) & 43\% / 1{,}092 / 15.7 & 51\% / 1{,}281 / 35.5 & 38\% / 582 / 14.8 & 8\% / $-3$ / 0.0 \\
CyberWorld (text) & 38\% / 886 / 19.7 & 35\% / 900 / 57.0 & 26\% / 192 / 72.2 & 26\% / $-2$ / 0.0 \\
CyberWorld (multimodal) & 14\% / 490 / 23.0 & 38\% / 1{,}186 / 0.5 & 28\% / 89 / 3.2 & 17\% / $-8$ / 0.0 \\
\bottomrule
\end{tabular}
\end{adjustbox}
\end{table}

%% file: appendices/appendix_C.tex
\section{Open-Loop Imagination of the Learned World Model}
\label{app:imagination}

\begin{figure}[h]
    \centering
    \includegraphics[width=\linewidth]{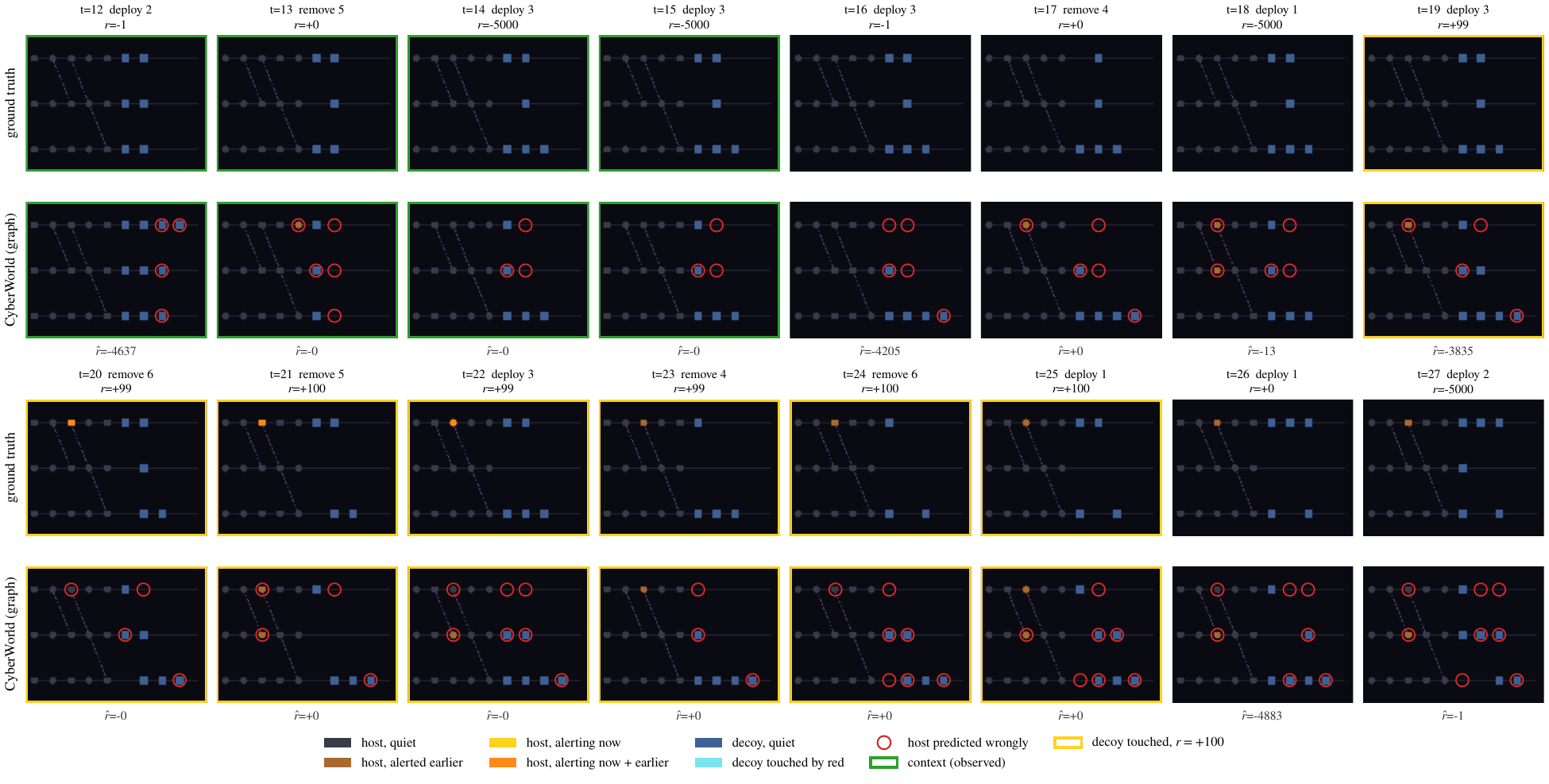}
    \caption{Open-loop imagination of the graph arm on a held-out BFSExfiltration
    episode. Top: the observed canvas at each step, annotated with the defender action
    that led to the step and the true reward. Bottom: the decoded model state, obtained
    from the posterior during the four observed steps (green border) and from the RSSM
    prior under the true defender actions thereafter, with the predicted reward shown
    below each canvas. Red rings mark hosts whose imagined state differs from the
    observed state; yellow borders mark steps on which the adversary touched a decoy.}
    \label{fig:imagination}
\end{figure}

\autoref{fig:imagination} visualizes open-loop trajectories imagined by the learned
world model. The model consumes the host representation and adjacency defined in
Section~\ref{sec:preprocessing}. The $192{\times}256$ semantic canvas shown here is
an exactly invertible rendering of the four reconstruction-role bits: hosts are drawn
as circles, decoys as squares, and node colour encodes the observable role state.
Because the rendering is invertible, decoded node states can be repainted onto the
same canvas as the observed states, making the two rows directly comparable.

The top row shows a held-out BFSExfiltration episode. The graph arm receives
observations for the first four displayed steps. After the fourth observed step, the
latent state is rolled forward open-loop using the true defender actions but without
further environment observations, and each imagined latent state is decoded and
repainted. Hosts whose imagined state differs from the observed state are marked with
red rings, while steps on which the adversary touched a decoy are marked with yellow
borders.

The topology is fixed by construction throughout the visualization, while the model
approximately preserves the observable host-state layer during the open-loop rollout.
Its errors are concentrated in two aspects of the decoded state. First, the model can
misplace deployed decoys among reserved slots within the same subnet. These slots are
implementation-level bindings that are only indirectly reflected in the observation,
so an imagined rollout may assign a deployed decoy to a neighbouring slot while
preserving the broader subnet-level structure. Second, the model can fail to anticipate
new alert activity. Once the rollout becomes fully open-loop, the prior receives no new
observation indicating that adversarial activity has reached another host, and the
decoded state can therefore continue to predict that host as quiet after the observed
trajectory begins to alert there.

The reward predictions expose the same limitation. Decoy-touch events are sparse and
depend on latent adversarial progress that is only partially revealed through subsequent
alerts. During the open-loop rollout, the predicted reward often remains near zero at
these events, and large negative rewards can be temporally misaligned with the observed
decoy-limit penalties. Thus, the figure should not be interpreted as demonstrating
pixel-perfect or event-perfect long-horizon prediction. Instead, it shows that the
learned latent dynamics preserve substantial coarse host-state structure while becoming
less reliable for events that depend on unobserved adversarial progress and precise
reward timing.

Despite these open-loop errors, the learned policy reaches the control in the
corresponding experiments. This suggests that precise long-horizon reconstruction of
every latent event is not necessary for acquiring the defensive behavior observed in
our evaluation, although improving prediction of adversary-dependent events remains an
important direction for future work.